%% file: Main.tex
\documentclass[11pt]{article}
\usepackage[a4paper]{geometry}
\usepackage{fullpage}
\usepackage{graphicx}
\usepackage{arydshln}
\usepackage{epsfig}
\usepackage{enumerate}
\usepackage{caption}
\usepackage{subcaption}
\usepackage{wrapfig}
\usepackage{float}
\usepackage[english]{babel}
\usepackage{mathrsfs}
\usepackage{amsmath}
\usepackage[nottoc]{tocbibind}
\usepackage{xcolor}
\usepackage{sectsty}
\usepackage{titling}
\usepackage{tfrupee}
\usepackage[T1]{fontenc}
\usepackage{titlesec}
\usepackage{blindtext}
\usepackage{parskip}
\usepackage{makeidx}
\usepackage{multirow}
\usepackage{etoolbox}
\usepackage{url}
\makeatletter
\providecommand{\subtitle}[1]{
  \apptocmd{\@title}{\par {\large #1 \par}}{}{}
}
\makeatother
\providecommand{\keywords}[1]{\textbf{\textit{Keywords---}} #1}
\title{Autoencoder based Hybrid Multi-Task Predictor Network for Daily Open-High-Low-Close Prices Prediction of Indian Stocks}
\author{Debasrita Chakraborty, Susmita Ghosh, Ashish Ghosh}
\date{}

\begin{document}

\maketitle
\abstract{Stock prices are highly volatile and sudden changes in trends are often very problematic for traditional forecasting models to handle. The standard Long Short Term Memory (LSTM) networks are regarded as the state-of-the-art models for such predictions. But, these models fail to handle sudden and drastic changes in the price trend. Moreover, there are some inherent constraints with the open, high, low and close (OHLC) prices of the stocks. Literature lacks the study on the inherent property of OHLC prices. We argue that predicting the OHLC prices for the next day is much more informative than predicting the trends of the stocks as the trend is mostly calculated using these OHLC prices only. The problem mainly is focused on Buy-Today Sell-Tomorrow (BTST) trading. In this regard, AEs when pre-trained with the stock prices, may be beneficial. A novel framework is proposed where a pre-trained encoder is cascaded in front of the multi-task predictor network. This hybrid network can leverage the power of a combination of networks and can both handle the OHLC constraints as well as capture any sudden drastic changes in the prices. It is seen that such a network is much more efficient at predicting stock prices. The experiments have been extended to recommend the most profitable and most overbought stocks on the next day. The model has been tested for multiple Indian companies and it is found that the recommendations from the proposed model have not resulted in a single loss for a test period of 300 days.}

\keywords{Stock Price Prediction, OHLC Data, Recommender System, Autoencoders, Long-Short Term Memory Networks, Multi-Task Learning, Hybrid Networks}

\input{SRS}

\bibliographystyle{plain}
\bibliography{Bibliography}
\end{document}

%% file: SRS.tex
\section{Introduction}
Stock price trends \cite{adam2016stock} is inherently very difficult to model due to its uncertain nature but has become one of the widely researched topics among the machine learning enthusiasts. Investors look forward to statisticians to predict the stock market trends. However, using traditional statistics, the prediction of prices becomes one of the most difficult tasks to do because of its dependency on multiple issues like physical factors, foreign markets trends, domestic market trends, news, people sentiments etc. This task is a multivariate time-series prediction problem. Such data is highly chaotic and can be treated as a complex adaptive system or dynamic system \cite{de2017complexity}. The problem is often tackled in two approaches. Researchers either look at market factors and try to predict the movement of stocks or study the historical to predict the future prices.

Thus, stock market analysis is broadly of two types: fundamental analysis \cite{chen2017enhancement, tiwari2012comparison} and technical analysis \cite{edwards2018technical}. The first one concentrates on the current economic and financial situation of the company, considering the company's assets, market value, debt inquired, profit, loss, balance sheet and other major economic reports describing the performance of the company. The second one studies the past market actions and price values, to determine its future performance. To develop machine learning models, the primary focus is usually on the price of the stocks, and the volume of stocks that are traded on a regular basis, and other factors include the supply and demand that captures the movement of the stock market.

In his theory, Burton \cite{malkiel2003efficient} points out that the stock market forecasting or modeling is not practical, as price fluctuations are unexpected in the actual world. Every movement in financial market pricing is dependent on current economic events or announcements. Traders are profit-oriented, and their judgments are taken based on most current occurrences, irrespective of prior analyses or expectations. The dispute about this hypothesis of an efficient market has never been resolved. There is no substantial evidence to show whether or not the efficient marketing theory is correct. But stock markets, according to Y. S. Abu-Mostafa \cite{abu1996introduction}, are to some extent predictable. The future finance markets are impacted by previous experience of numerous price fluctuations over a period in the stock sector and from the unreported serial connections of critical economic situations.

Stock prices from history are used to predict the future value of a stock corresponding to the stock exchange. These predicted stock prices helps the investors or traders to decide whether they want to trade stocks now or hold them for some time. Therefore, stock market analysis can be broadly classified into, short term trading, high-frequency trading, long-term trading, intraday trading \cite{tsai2019assessing} and interday trading \cite{karalevicius2018using}. This work concentrates on buy-today and sell-tomorrow (BTST) trading only. Traders buy and sell stocks in large numbers with the intention of booking the highest amount of profit in a day. Such trading, being short term, is a lot more riskier than investing in the regular stock market. So, it is very crucial to know the prices of the stocks on the next day to gain profit. Various models \cite{ballings2015evaluating, singh2017stock, kim2018forecasting} have been proposed in this regard to predict the prices of stocks. However, almost all the models try to predict the movement of the prices using some indicator variables \cite{colby2003encyclopedia} which are mostly modelled on the opening, high, lowest and closing (OHLC) prices of a stock on a particular day. Studies on direct prediction of OHLC prices are not so widespread in the machine learning community. Although there have been some attempts to predict the closing prices \cite{budiharto2021data}, high prices \cite{gorenc2016prediction} or the average prices \cite{manurung2018algorithm} independently, the literature has somewhat ignored the possibility of modelling OHLC prices together.

With the advancement in the field of neural networks, it has become comparatively easier to model the stock prices variations up to a fairly good extent \cite{zahedi2015application}. Long Short Term Memory (LSTM) networks \cite{greff2016lstm} and Gated Recurrent Unit (GRU) networks \cite{wu2020adaptive} have become the state of the art models for efficiently predicting the stocks. But, these neural networks come with a cost. If the normalisation of the input variables are not done carefully, the predictions often end up being faulty \cite{panigrahi2013effect}. This calls for a new variation in normalisation of stock prices. Moreover, since there is an angle of financial risk to these models, they need to be as accurate as possible in their predictive capabilities. As argued by most deep learning researchers \cite{hussain2019cloud}, recurrent neural networks (RNNs) perform better predictions if the raw data is transformed to a different feature space. There have been studies \cite{chakraborty2019integration} which show that autoencoders (AEs) are good feature extractors. This has been the motivation consider such AE based feature extracting methods for predicting stocks. Combined with the time series forecasting abilities of LSTM networks and feature learning capabilities of AEs, this manuscript presents a hybrid network which takes help of multi-task learning to learn the OHLC relationships. It is seen that using the proposed model, the most profitable stocks can be recommended efficiently. A momentum indicator known as William's \%R \cite{edwards2018technical} is used to determine the stocks which would be overbought the next day. Such a recommender system would help the investors make profitable decisions while trading.

Existing research on predicting open, high, low and close price (OHLC) data suffers from two flaws. There are two main reasons behind this. This means that all values must be positive, the low price must always be less than or equal to the high price, and that both the open and close prices must fall between two defined limits in order for the OHLC data to be valid. Constraints must be taken into account when using the time-series modelling techniques for the four variables of OHLC data. The contributions of this manuscript is three fold, namely,
\begin{enumerate}[(i)]
\item A novel normalisation approach for the prediction of OHLC prices,
\item a novel hybrid multi-task network of AE and LSTM-RNN, and
\item a system to recommend the most profitable and overbought stocks (using William's \%R indicators) for the next day.
\end{enumerate}

The remaining manuscript gives a detailed overview of the work presented. The following section \ref{C5sec2} contains the description of related works. Section \ref{C5sec3} gives a preliminary description of the building blocks of the proposed method. Section \ref{C5sec4} contains the dataset description, data pre-processing, and describes the working of the system. In section \ref{C5sec5}, experimental results are shown containing all the result visualization. The conclusions drawn from the proposed system has been provided at the end in section \ref{C5sec6} followed by the scope of future works.

\section{Related Works}\label{C5sec2}
A commendable amount of work has been done in the field of stock price prediction. Past studies \cite{fama1965behavior} have shown that it is impossible to predict the stock prices accurately as it depends on so many factors including the domestic market trends, population sentiments, news, foreign market trend, etc. The stock price time series follows a random walk model which makes it less likely to be predictable. However, with the recent advancements in the field of machine learning, it has become possible to predict the stock prices accurately up to a fairly good extent.

There has been a benchmark survey \cite{ballings2015evaluating} of techniques including Logistic Regression, Support Vector Machines (SVM), Artificial Neural Networks (ANN), $k$-Nearest Neighbor ($k$NN) and other ensemble models, which can predict stock prices. Researchers \cite{nayak2015naive} implemented SVM along with $k$NN to analyse stock market trends on Indian stocks. Researchers \cite{mondal2014study, adebiyi2014comparison} have been much appreciative of autoregressive models like ARIMA (auto-regressive integrated moving average) which have been proven to be robust for short-term forecasting. A triple stage method was proposed \cite{enke2013stock} for forecasting the stock prices. It included a stepwise regression analysis, grouping of selected variables by a differential evolution-based fuzzy clustering method and a fuzzy inference neural network for final prediction. A similar work \cite{zhong2017forecasting} has also been proposed which combines dimensionality reduction using principal component analysis (PCA) with ANN for S\&P 500 Index exchange-traded fund (SPY) stock market returns. Hybrid models of fuzzy time series and automata \cite{talarposhti2016stock}, granular computing \cite{chen2015hybrid} were also explored. Support vector regression (SVR) models have been explored \cite{henrique2018stock} for stock price prediction. Hybrid of neural networks \cite{wang2019stock} have also been used to predict the volatility of stocks.

Deep neural networks are the state of the art models \cite{singh2017stock} for stock prediction. Recently, research has been diverted to using LSTM \cite{hochreiter1997long, greff2016lstm, fischer2018deep} models. Researchers \cite{kim2018forecasting} have used LSTM and GARCH \cite{francq2019garch} to forecast the volatility of stock price index. A standard LSTM network \cite{hochreiter1997long, fischer2018deep} or a standard GRU network \cite{shen2018deep} would simply takes in the stock prices and predicts the next day's price. LSTM networks have been used to predict the closing price of the stocks \cite{budiharto2021data} recently. However, there are only a handful of works to the knowledge of the authors which have handled the prediction of OHLC prices. It may be argued that predicting OHLC prices directly makes more sense as the technical indicators used for stock price modelling are built on these OHLC values only and a novice trader will not understand the meaning of these indicators. However, anyone can understand that one can make high profit is he buys a stock closer to the low price and sells it on the highest price. The bullish and bearish signals can also be judged by looking at the opening and closing price for the next day. If the opening price is greater than the closing price, it is a bearish signal and indicates that the market is going to be unfavourable. On the other hand if the opening price is lesser than the closing price, it is a bullish signal and indicates that the market is going to be favourable. Thus, the present study concentrates on efficient prediction of OHLC prices directly.

A novel loss function based LSTM network has been proposed to predict the opening, high, lowest and closing prices separately known as FLF-LSTM \cite{ahmed2020flf}. Even though they have not explicitly identified or handled the OHLC constraints, it is found that the model is capable of making meaningful OHLC predictions. Another research \cite{manurung2018algorithm} proposed predicting the average of the OHLC using an LSTM network, but was eventually unable to make meaningful predictions as seen in the experiments. A bi-directional LSTM based multi-task model has been proposed to predict OHLC prices \cite{mootha2020stock}. It is shown \cite{baxter1997bayesian} that multi-task models which make use of hard parameter sharing \cite{reyes2019performing} are less prone to overfit and can efficiently learn the relationships in the data. However, the proposed model fails to make any meaningful predictions and the predicted values do not follow the constraints of an OHLC data (explained in section \ref{c5ohlc}). This is mainly because their model assumes independence of the OHLC values in their proposed model as they start with the task-specific layers in the initial layers and then try to aggregate using a shared layer in the latter part of the network. Thus, the proposed work aims to take a critical angle at predicting meaningful OHLC prices and also reducing the financial losses incurred by following the less accurate stock price prediction models. The following section explains some preliminary concepts used in the proposed method.

\section{Preliminaries}\label{C5sec3}
Stock market predictions are used to estimate the true value of stock for the upcoming days. Basically, the aim is to understand the future behaviour of the stock market on the basis of previous performance of the stocks and the information about the company's stocks.

To analyze the stocks is a very challenging task, as it requires a deep understanding of the stocks and market trends, and of other factors, including current industrialized policies, effect of foreign markets, country's financial situation, past market behavior, etc. The other major difficulty is the volatility of the stock market \cite{adam2016stock, engle2013stock} which makes it nearly impossible to understand the trends in the stock price. With the increasing research in the field of RNNs, such time varying trends are easily recognised. The following section gives a very brief overview of the LSTM cell and the OHLC properties.

\subsection{Long Short Term Memory (LSTM)}
Long Short Term Memory (LSTM) \cite{zhao2017lstm} cell is a special kind of RNN node with a forget gate. A LSTM cell consist of four parts i.e. input gate ($\textbf{i}_t$), output gate ($\textbf{o}_t$), cell state ($\bar{\textbf{C}}_t$), and forget gate ($\textbf{f}_t$). The LSTM cell\cite{greff2016lstm} takes three inputs, the output from previous unit $\textbf{h}_{t-1}$, the memory of the previous unit (cell state) $\textbf{C}_{t-1}$, and the input of the current time step $\textbf{x}_{t}$.

\begin{figure}[htp]
    \centering
    \includegraphics[width=0.6\linewidth]{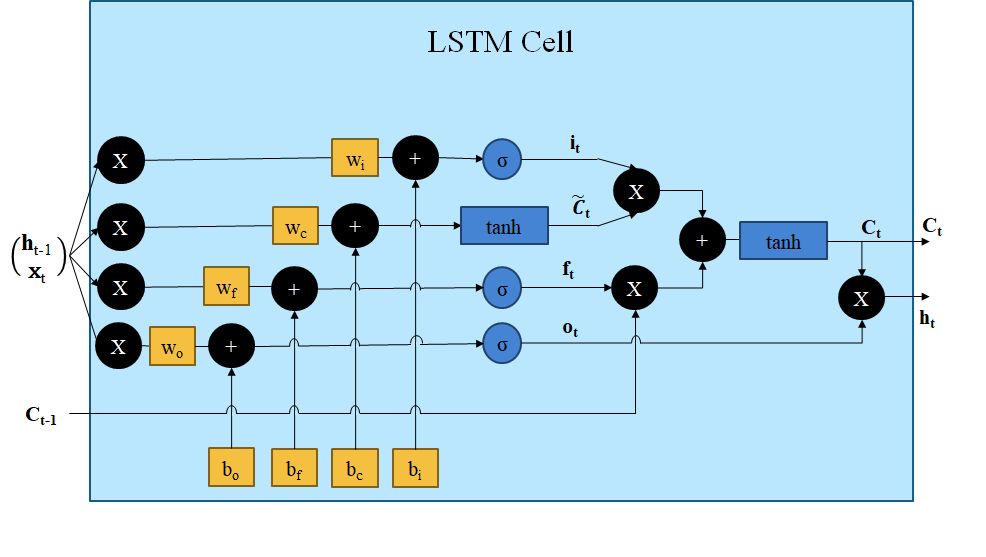}
    \caption{A pictorial representation of LSTM Cell.}
    \label{fig:LSTM node}
\end{figure}

This single unit makes decision by considering the three features and produces a new output, and alter its memory. $\textbf{C}_{t}$ and $\textbf{C}_{t-1}$ represents the cell state at time step $t$ and $t-1$ respectively, similarly $\textbf{h}_{t}$ and $\textbf{h}_{t-1}$ represents the hidden state output at time step $t$ and $t-1$ respectively. $\textbf{x}_{t}$ is the input at time $t$ [Figure \ref{fig:LSTM node}]. The input $\textbf{x}_{t}$ and the previous hidden state output $\textbf{h}_{t-1}$ are concatenated as $[\textbf{h}_{t-1},\textbf{x}_{t}]$ and are multiplied by respective weights for each gate.

Forget gate takes two inputs i.e. $\textbf{h}_{t-1}$ and $\textbf{x}_{t}$, representing the output at time $t-1$, and input at time $t$, respectively. The vectors $\textbf{h}_{t-1}$ and $\textbf{x}_{t}$ are concatenated to form a vector $[\textbf{h}_{t-1},\textbf{x}_{t}]$ and is multiplied by the forget weight matrix $\textbf{W}_{f}$. The equation governing this operation is given by equation \ref{eqn2}.

\begin{equation}\label{eqn2}
     \textbf{f}_t=\sigma(\textbf{W}_f.[\textbf{h}_{t-1},\textbf{x}_{t}]+\textbf{b}_f).
\end{equation}

Here, $\textbf{f}_t$ represents the forget gate output and $\textbf{b}_f$ is the forget gate bias. The forget gate, decides the amount of information to retain and it will forget the remaining information.

Next, the input gate decides the amount of information addition to the cell state. The concatenated vector $[\textbf{h}_{t-1},\textbf{x}_{t}]$ is passed to the sigmoid function, and the tanh activation function simultaneously. The sigmoid gate represents the input gate [equation \ref{eqn3}] and the tanh gate represents the cell state gate [equation \ref{eqn4}].

\begin{equation}\label{eqn3}
     \textbf{i}_t=\sigma(\textbf{W}_i.[\textbf{h}_{t-1},\textbf{x}_{t}]+\textbf{b}_i),
\end{equation}

where, $\textbf{W}_i$ represents the input gate weights, $\textbf{i}_t$ represents the input gate output and $\textbf{b}_i$ is the input gate bias. The input gate, decides the amount of information that would be updated.

\begin{equation}\label{eqn4}
     \bar{\textbf{C}}_t=tanh(\textbf{W}_C.[\textbf{h}_{t-1},\textbf{x}_{t}]+\textbf{b}_C).
\end{equation}

Here, $\textbf{W}_C$ represents the cell state gate weights, $\bar{\textbf{C}}_t$ represents the candidate cell state gate vector and $\textbf{b}_C$ is the cell state gate bias. In the cell state gate, a vector of new candidate values $\bar{\textbf{C}}_t$ are created, which would be added to the final cell state.

The final cell state is given by equation \ref{eqn5}.

\begin{equation}\label{eqn5}
     \textbf{C}_t=\textbf{f}_t.\textbf{C}_{t-1}+\textbf{i}_t.\bar{\textbf{C}}_t.
\end{equation}

The cell state transfers relevant information down the sequence processing. Thus, it can learn long-term dependencies. Now, the concern is about which filtered part of the cell state is needed to go to the output. It is decided by the output gate result $\textbf{o}_t$, given by,

\begin{equation}\label{eqn6}
     \textbf{o}_t=\sigma(\textbf{W}_o.[\textbf{h}_{t-1},\textbf{x}_{t}]+\textbf{b}_o).
\end{equation}

Here, $\textbf{W}_o$ represents the output gate weights and $\textbf{b}_o$ is the output gate bias.

The final hidden state output thus can be written as,

\begin{equation}\label{eqn7}
     \textbf{h}_t=\textbf{o}_t.tanh(\textbf{C}_{t}).
\end{equation}

Using these four gates, the LSTM can decide which information to forget and which information to retain down the memory and thus can be used to model time-series data efficiently.
\subsection{Open-High-Low-Close (OHLC) Constraints}\label{c5ohlc}
Daily stock prices are often represented in terms of four variables namely opening, high, low and closing prices. The names of these variables indicate straightforward price values. Since the stock prices vary in real-time, the data of a day is stored with respect to how high the stock price has been (high price), how low it has been (low price), at what price of the stock the market opened (open price) and at what price of the stock the market closed (close price). The open price of a particular stock is determined by the demand and supply at the start of the market and close price is the final price at which the stock is sold for the day. The Bombay Stock Exchange usually opens at 9.15 a.m. and closes at 3.30 p.m. Thus, the maximum value that the stock has attained within this time is called the high price and the minimum value that the stock has attained within this time is called the low price.

Thus, there are some clear indications as to what kind of constraints these variables should have. Then, it is obvious that the high price will have the highest value and the low price will the lowest value among these four variables. The open and close price should lie somewhere in between the former two extremal values. Let us denote the open, high, low and close prices of a particular stock as $X_O$, $X_H$, $X_L$ and $X_C$. The following constraints exist which should be maintained during prediction tasks.
\begin{enumerate}[(i)]
\item{The values of the prices must not be negative or zero.
\begin{eqnarray}\label{ohlc1}
X_O& >& 0,\\
X_H& >& 0,\\
X_L& >& 0, and\\
X_C& >& 0.
\end{eqnarray}}
\item{The high price should be greater than the low price.
\begin{equation}\label{ohlc2}
X_H>X_L
\end{equation}
}
\item{The open and close price should lie in between the high and the low price.
\begin{eqnarray}\label{ohlc3}
X_O &\in& [X_H,X_L], and\\
X_C &\in& [X_H,X_L].
\end{eqnarray}
}
\end{enumerate}

\begin{figure}[htp]
    \centering
    \includegraphics[width=0.6\linewidth]{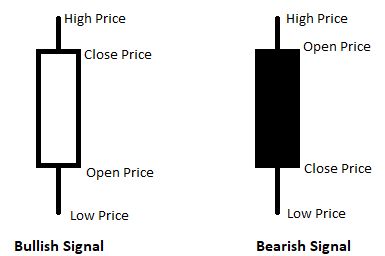}
    \caption{Pictorial representation of Japanese candlestick chart for OHLC data.}
    \label{fig:JC}
\end{figure}

These are also often represented as a Japanese candlestick chart \cite{meng2018research} [Figure \ref{fig:JC}].

When the body is solid or filed in black, it indicates a bearish signal and when it is hollow or white, it indicates a bullish signal. There are many interesting patterns which can be found in such OHLC based charts and often depict emotions of the people surrounding a stock and are hence much informative than other trading indicators.
\subsection{Multi-task Learning}
\begin{figure}[htp]
    \centering
    \includegraphics[width=0.6\linewidth]{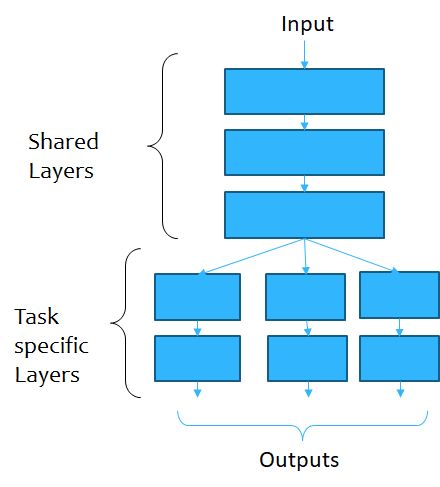}
    \caption{Pictorial representation of an multi-task learning model with shared and task-specific layers.}
    \label{fig:MTL}
\end{figure}

While we are training a ML model for a certain task, the traditional method is to train a single model or an ensemble till its performance no longer improves. This makes the model to focussed on that specific task and so, we often ignore how this model can be utilised to learn for some related tasks. Transfer Learning \cite{pan2009survey} has been explored in the attempt to learn related tasks. However, such tasks are often independent of each other and mostly belong to different domains. So, they can utilise the same optimisation function for both the tasks. Multi-task learning \cite{mootha2020stock, zhang2018overview} (MTL) enables sharing of representations which takes into account the relationships between the related tasks, even if they are constrained. MTL models have two main types of layers when it comes to neural networks: shared layers and task specific layers [Figure \ref{fig:MTL}].
The shared layers perform a sort of inductive transfer of information across the multiple tasks and the task-specific layers allow the independent learning of tasks. In the present case, we have the tasks of predicting the OHLC values which may seem to behave independently on a macro level (as the day's opening price cannot judge what the lowest or highest will be), but are constrained in a micro level (as described in section \ref{c5ohlc}). MTL allows a restrained independence by drawing insights from it. This particular property allows an MTL model to be prone to overfitting as it is trying to capture the representations of all the tasks simultaneously. It has been shown \cite{baxter1997bayesian} that as the number of tasks grows, the risk of overfitting the shared parameters becomes less than the risk of overfitting the task-specific layers.
\section{Proposed Method}\label{C5sec4}
The proposed method takes advantage of the feature extracting capabilities of the autoencoder and the predictive abilities of LSTM network. The experiments show that using a single-task learning model to predict OHLC prices is not enough to have profitable recommendations for investable stocks. So, a MTL model is proposed in this regard. The entire method deals with prediction of OHLC prices of stocks and then recommending the relevant stocks to the user. It is observed that the proposed method does not result in a single loss over a test period of 350 days if the top profitable recommended stocks are actually invested in. The following segments describe the datasets and the proposed method in detail.
\subsection{Dataset Description}
In this analysis twenty seven Indian companies' stocks are considered which are listed by Fortune India 500. The data has been taken from the Bombay Stock Exchange website (\url{www.bseindia.com}). The datasets contain the open, high, low and close prices of the 27 stocks from the date 01-01-2014 to date 01-01-2019. There are several days in between this date range when the stock market was closed due to holidays. So, the effective number of samples were only available for 1234 days. The last 350 days (about 30\%) from 02-08-2017 to 01-01-2019 for all the companies are taken as the testing period and the former days (from 01-01-2014 to 01-08-2017) are taken for training (around 70\%) and validation of the model.
\subsection{Proposed Normalisation Strategy}
The main restriction to stock-price modelling lies in the normalisation strategy. For any neural network, a proper normalisation always offers benefits \cite{sola1997importance}. It has been show \cite{bhanja2019impact} that since stock prices have a wide range of values, normalisation of the prices to a fixed scale improves the prediction capabilities of deep RNNs. In this context, several normalisation methods have been used in the literature. Since the prediction models often take a window of past values into account to predict the next day's value, most of the methods till rely on the min-max normalisation of the prices over that window. For example, if a stock is studied for 100 days, its price may fluctuate from the order of tens to the order of thousands within a few days. If the window is chosen to be 20, then one would get 81 overlapping windows of data each of which will have a desired prediction value for training. And each of these 81 windows will be individually normalised using min-max normalisation. So, the price value which is the highest among each window will be mapped to 1 and the lowest will be mapped to zero. This implies that the same value may get different mappings if the window contains different values. For training data, this normalisation is easy. But, for data where the desired prediction value is not present, or for days when the prediction value makes a sudden jump from a very low range to a very high range or from a high range to a very low range, the predicted value should ideally be more than 1 or less than 0 respectively. In that case, de-normalisation becomes faulty. So, a proper normalisation strategy is important which will handle all the situations and all possible values of prices. A sigmoid-like function is much desirable in such cases where the values $[-\infty, \infty]$ is mapped to $[0, 1]$. However, using sigmoid straightaway is again problematic as any value greater than two will be mapped near to 1. So, the values in the order of hundreds or thousands will not be as much differentiable from one another. Now, one may utilise the properties of OHLC prices (that they cannot be negative or zero), and construct a logarithmic alternative. Thus, the proposed normalisation strategy is given as,
\begin{equation}
y'=\dfrac{1}{(1+e^{-z})},
\end{equation}
where,
\begin{equation}
z=log_{10}(1+x).
\end{equation}
Then, the de-normalisation is straightforward given by,
\begin{equation}
x=10^{(-ln(\dfrac{1}{y'}-1)}-1,
\end{equation}
The factor 1 is added to $x$ to avoid any computational restrictions for very low (>0) values of $x$. This normalisation is independent of any window operation and may be applied to any unknown price ranges as well. It can be proven that if the proposed normalisation is applied to OHLC data, the normalised values are also OHLC in nature. The normalised values lie in the range $(0.5,1]$ for $x$ in range $(0,1]$. Thus, for OHLC, the values $X_O$, $X_H$, $X_L$ and $X_C$ being greater than zero, the normalise values are also greater than zero. Moreover, since the sigmoid function and the logarithmic function is monotonically increasing for $x$>0, then the other constraints of OHLC also hold which makes the normalised value of $X_H$ the highest and the normalised value of $X_L$ the lowest. The normalised values of $X_C$ and $X_O$ will lie between the normalised value of $X_H$ and $X_L$ respectively. Let us denote the normalised value of $X_O$, $X_H$, $X_L$ and $X_C$ as $y'_O$, $y'_H$, $y'_L$ and $y'_C$. These normalised values capture the variation of price ranges with respect to time. However, we need one extra set of features from the OHLC data, which would capture the variation of the prices with respect to each other without any temporal component to them.

We can construct another normalisation such that for prices $X_O$, $X_H$, $X_L$ and $X_C$, the normalised prices are given by,
\begin{eqnarray}
  y_O &=& \dfrac{X_O-X_L}{X_H-X_L}, \\
  y_H &=& \dfrac{X_H-X_L}{X_H}, \\
  y_L &=& \epsilon, and \\
  y_C &=& \dfrac{X_C-X_L}{X_H-X_L}.
\end{eqnarray}

The values $y_O$, $y_H$, $y_L$ and $y_C$ all lie in the range $(0,1]$. This is equivalent to row-wise normalisation of the variables, which incorporate inter-variable relationships.

Thus, the proposed method uses eight variables $\{y’_O, y’_H, y’_L, y’_C, y_O, y_H, y_L, y_C\}$ for each date as input to the model.
\subsection{Model Architecture}
\begin{figure}[htp]
    \centering
    \includegraphics[width=0.6\linewidth]{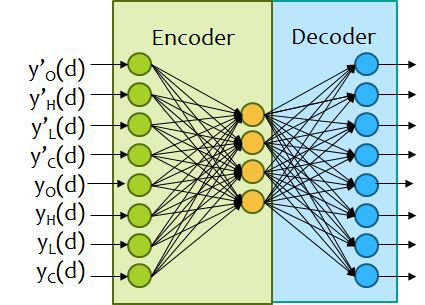}
    \caption{Pictorial representation of the AE network used in the proposed method.}
    \label{fig:PAE}
\end{figure}

\begin{figure}[htp]
    \centering
    \includegraphics[width=0.9\linewidth]{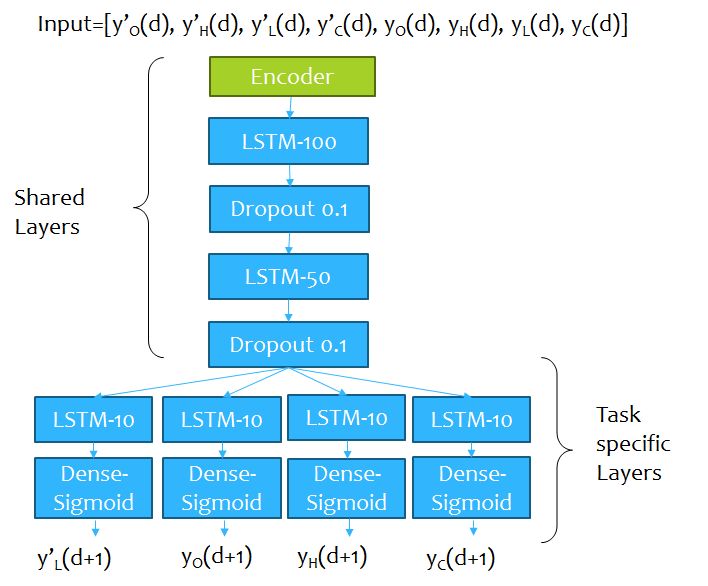}
    \caption{Pictorial representation of the proposed multi-task learning model.}
    \label{fig:PMTL}
\end{figure}

The proposed model consists of an MTL model consisting of LSTM layers preceded by an autoencoder's encoder pre-trained on the prices of the training days to reconstruct the input. The encoder of the pre-trained autoencoder is cascaded before the input of the proposed MTL model. There was a single hidden layer in the autoencoder with 4 hidden units. The number of hidden units for the AE network was selected using Shibata Ikeda's method \cite{shibata2009effect}. The architecture of the AE used is given in Figure \ref{fig:PAE}.
After the AE is trained, the decoder is discarded and the encoder is used to extract the features from the eight variables of each date. These features are then fed to the predictor network for further prediction. The predictor network is constructed in a MTL fashion having some shared and some task-specific layers. It should be noted that the encoder is also a part of the shared layers if the entire model is considered. A pictorial representation of the entire model is given in Figure \ref{fig:PMTL}. The choice of the number of shared layers and task specific layers have been done using cross-validation and is elaborated in section \ref{C5sec5}.
Since the predictions need to be made on the next day's OHLC values, one may predict any one of $y’_O$, $y’_H$, $y’_L$ and $y’_C$ to get the range of values and rest three of $y_O$, $y_H$, $y_L$ and $y_C$ for finding the inter-variable relationship for the next day. The last dense layers consists of sigmoid activation function. Thus, the output prediction is always in the range $[0,1]$ as desired. In the proposed model, the four variables chosen to be predicted are $y'_L$, $y_O$, $y_H$ and $y_C$. Let the predicted values of the four variables are $y'^{Pred}_L$, $y^{Pred}_O$, $y^{Pred}_H$ and $y^{Pred}_C$. It was seen by experiments (section \ref{C5sec5}) that the low price is less volatile than all the other prices. So, predicting the range of values based on the low price would be much more suitable and other price values may be rebuilt using that. Thus, the actual predicted values of prices may be rebuilt using the following equations,
\begin{eqnarray}
  X_L^{Pred} &=& 10^{(-ln(\dfrac{1}{y'^{Pred}_L}-1)}-1, \\
  X_H^{Pred} &=& \dfrac{(1-y_H^{Pred})}{X_L^{Pred}} \\
  X_O^{Pred} &=& y_O^{Pred}.(X_H^{Pred}-X_L^{Pred})+X_L^{Pred} \\
  X_C^{Pred} &=& y_C^{Pred}.(X_H^{Pred}-X_L^{Pred})+X_L^{Pred}
\end{eqnarray}

Thus, by default these predicted values maintain the OHLC constraints.

The model is trained using a sliding window method \cite{xie2020bioacoustic}. A look-back period or window size of 20 days is taken to predict the next day's prices. For the training data, the predicted day's values are matched with the actual values for that day and the model is trained to reduce the mean squared error of each price independently in the task specific layers. For the testing days, the past 20 days' values are used to predict the next day's values.

\subsection{Recommendations}\label{rec}
After the training, the Opening Price, Closing Price, High Price and Low Price for the next day are predicted for each stock, taking the last 20 days' features as input. Since the proposed method considers only intra-day trading scheme, we are only concerned about the behaviour of the stocks on the next day. As the behaviour is predicted one day ahead, the stocks that the user may buy the next day can be efficiently recommended. Using the predicted data for the $i^th$ day, two indicator variables namely Profit and William's \%R \cite{dash2016hybrid} are constructed for the $i^th$ predicted day.

Let us use the abbreviations where, $X_O^{Pred}(i)$ indicate the predicted opening price, $X_H^{Pred}(i)$ indicate the predicted highest price, $X_L^{Pred}(i)$ indicate the predicted lowest price and $X_C^{Pred}(i)$ indicate the predicted closing price on the $i^{th}$ day. Similarly, $X_O^{Actual}(i)$ indicate the actual opening price, $X_H^{Actual}(i)$ indicate the actual highest price, $X_L^{Actual}(i)$ indicate the actual lowest price and $X_C^{Actual}(i)$ indicate the actual closing price on the $i^{th}$ day.

It is quite intuitive that the maximum profit one can attain in a particular stock is when he/ she buys the stock on the previous day $i-1$ (when the price was the lowest i.e. at $X_L^{Actual}(i-1)$) and holds it for the next day $i$ till it achieves a highest price ($X_H^{Actual}(i)$) will be given as equation \ref{eqn10}.

\begin{equation}\label{eqn10}
     Actual Profit ={(X_H^{Actual}(i) - X_L^{Actual}(i-1))},
\end{equation}

So, one may sell closer to the predicted highest price on $i^{th}$ day to attain a profit close to the actual profit. The profit obtained by the person by following the proposed model's recommendation is given by equation \ref{eqn101}.

\begin{equation}\label{eqn101}
     Profit ={(X_H^{Pred}(i) - X_L^{Actual}(i-1))},
\end{equation}

Another important momentum indicator used by traders is the William's \%R. It is used to know about the stocks which a user should buy. Its value oscillates between values 0 to -100. It is given as equation \ref{eqn11}.
\begin{equation}\label{eqn11}
     William's \%R =\dfrac{(X_H(Max) - X_C^{Pred}(i)}{(X_H(Max) - X_L(Min))}\times 100\%,
\end{equation}
Here, $X_H(Max)$ signifies the highest high price obtained in the recent $d$ days. The value of $d$ is usually chosen to be 14 by the stock analysts. In order to choose the $X_H(Max)$, the past $d-1$ days of actual high price data is present. So, including the predicted high price $X_H^{Pred}(i)$ in consideration would make more sense to calculate the predicted value of William's \%R. So, $X_H(Max)$ is given as,
\begin{multline}\label{eqn15}
X_H(Max)=max(X_H^{Actual}(i-d), X_H^{Actual}(i-d+1), X_H^{Actual}(i-d+2),...\\ X_H^{Actual}(i-1),X_H^{Pred}(i))
\end{multline}

Also, $X_L(Min)$ signifies the lowest low price achieved in the recent $d$ days. Since the actual data for $d-1$ days is present and one can only predict the lowest price of the next day ($X_L^{Pred}(i)$), $X_L(Min)$ is calculated as,

\begin{multline}\label{eqn16}
X_L(Min)=min(X_L^{Actual}(i-d), X_L^{Actual}(i-d+1), X_L^{Actual}(i-d+2),... \\ X_L^{Actual}(i-1),X_L^{Pred}(i)).
\end{multline}

It is indicated that a William \%R mark below -80 is considered to be an oversold case \cite{edwards2018technical} and suggests a buying signal.

In order to test the proposed model's efficacy, we are recommending stocks in two phases:

\begin{enumerate}[(i)]
\item Highest profitable stocks: From the list of the profits generated by each stock on a particular day, the top company is chosen, which can maximize the predicted amount of profit on the next day.

\item Stocks which will be overbought: The predicted William's \%R is calculated and if any stock crosses the -80 mark (which suggests an overbought case), it is recommended.
\end{enumerate}

The following section describes the various experiments conducted and results obtained for all the 27 Indian companies' stocks.
\section{Experimental results}\label{C5sec5}
There are several aspects of experimentation as the proposed method tries to handle the problem of stock price prediction from various perspectives. The first experiment reveals why the proposed normalised low price ($y'^{Pred}_L$) was chosen to be used as a prediction variable. The second experiment looks into various existing models and checks whether those really make meaningful predictions or not. The third experiment shows a comparison of various methods
\subsection{Low Price is Less Volatile}
\begin{figure}[htp]
    \centering
    \includegraphics[width=0.8\linewidth]{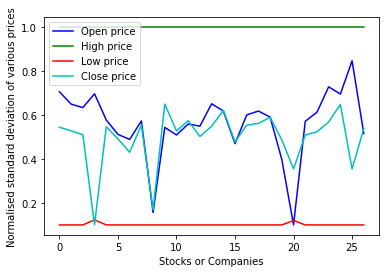}
    \caption{Graph showing the volatility of different prices.}
    \label{fig:LowVol}
\end{figure}

It was shown by Gorenc and Velu{\v{s}}{\v{c}}ek \cite{gorenc2016prediction}, that high price is less volatile than closing price. However, their experiments did not take the low and opening price into their consideration. Following the same argument, if we plot the normalised standard deviations of the various prices of the companies for the entire training period data, it is seen that the low prices is the least volatile [Figure \ref{fig:LowVol}] for most companies.

The standard deviation is normalised to a range 0 to 1. Normalised standard deviation has been used in this case because the different companies had different price ranges and required mapping to a same range to view the effective results. It can be seen that the high prices have the highest volatility (highest standard deviation), which contradicts the earlier findings. This is mainly due to the fact that the authors used a window based normalisation of OHLC features prior to calculating the standard deviation which contributed to the faulty interpretation.
\subsection{How do Autoencoders Benefit in Stock Data?}
\begin{figure}[htp]
    \centering
    \includegraphics[width=0.8\linewidth]{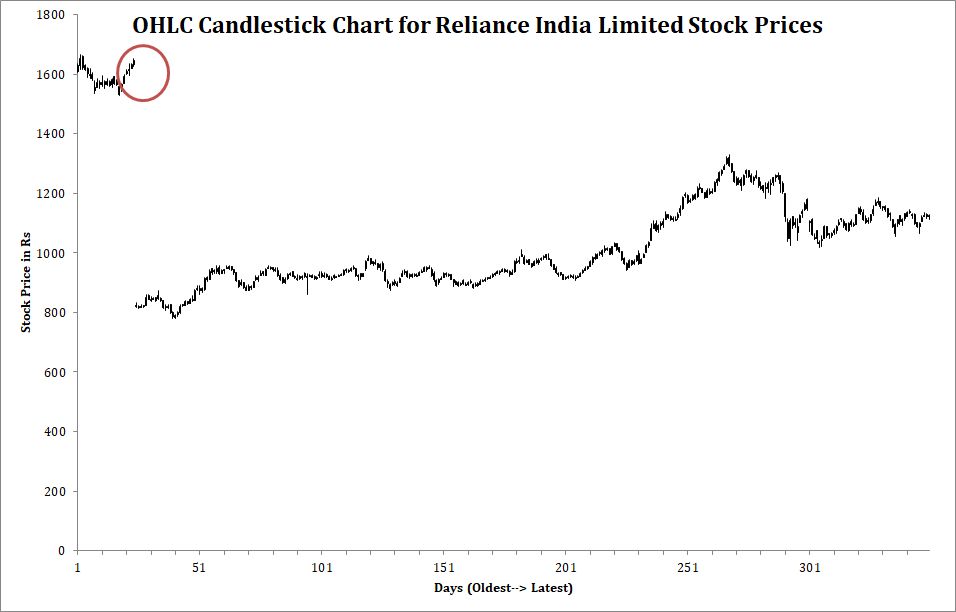}
    \caption{OHLC candlestick chart for Reliance India Limited for the test days. (The point of drastic decrease in price has been marked by red in the figure.)}
    \label{fig:ICICIOHLC}
\end{figure}

AEs have a tendency to map similar data to similar ranges. Now, if a particular stock has drastic value fluctuations within a very short period of time, the actual OHLC feature space would be overwhelmed with the disparity in values and will fail to capture the interesting trends in the data. This is analogous to a sudden event in the time. However, it cannot be directly termed as an anomaly as the state of increased or decreased value persist for a long time after it has started. This has been observed for many stocks in the dataset. One such stock is of Reliance India Limited, where in the test days, the value of the stock fell drastically (to almost half the value in a single day) as seen in Figure \ref{fig:ICICIOHLC} (marked in red).

\begin{figure}[htp]
    \centering
    \begin{subfigure}{0.45\textwidth}
    \centering
    \includegraphics[width=1\linewidth]{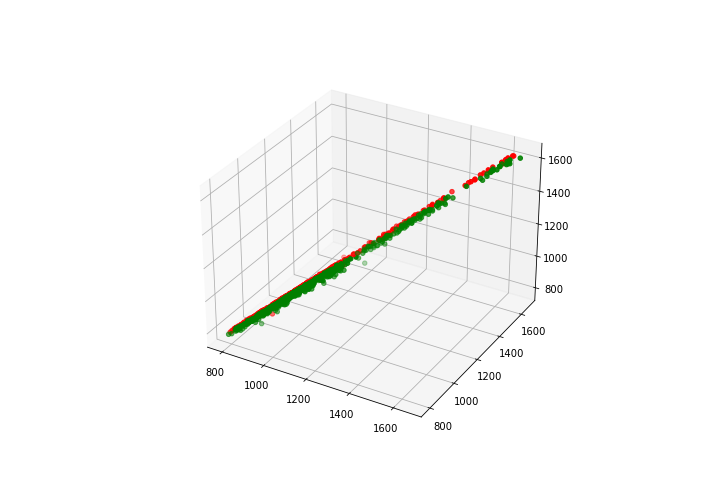}
    \caption{}
    \end{subfigure}
    \begin{subfigure}{0.45\textwidth}
    \centering
    \includegraphics[width=1\linewidth]{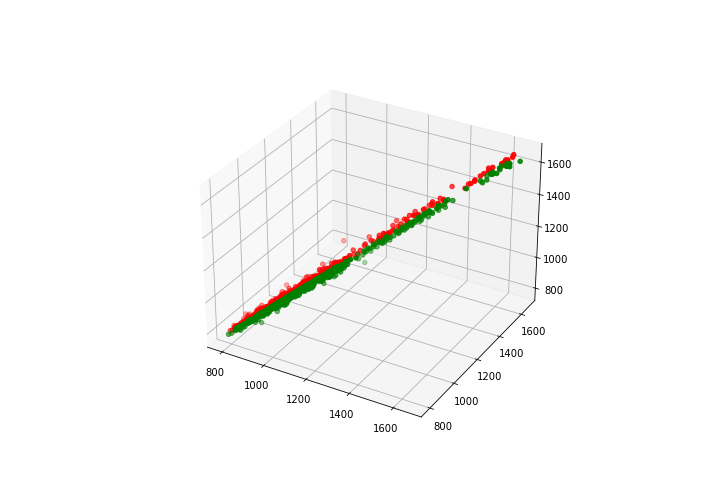}
    \caption{}
    \end{subfigure}
    \begin{subfigure}{0.45\textwidth}
    \centering
    \includegraphics[width=1\linewidth]{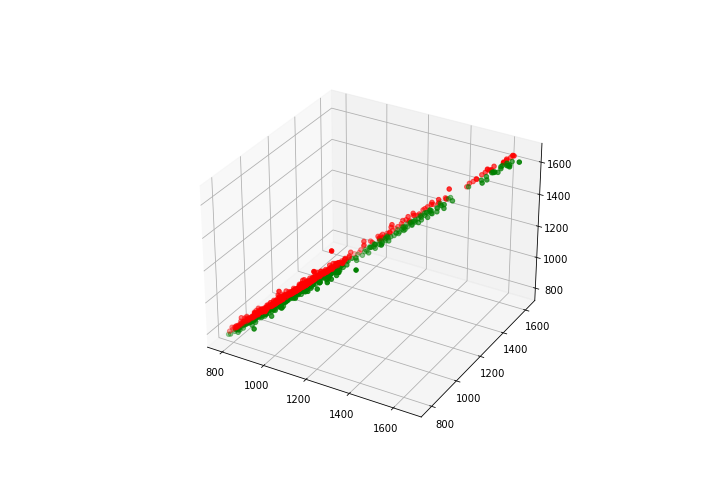}
    \caption{}
    \end{subfigure}
    \begin{subfigure}{0.45\textwidth}
    \centering
    \includegraphics[width=1\linewidth]{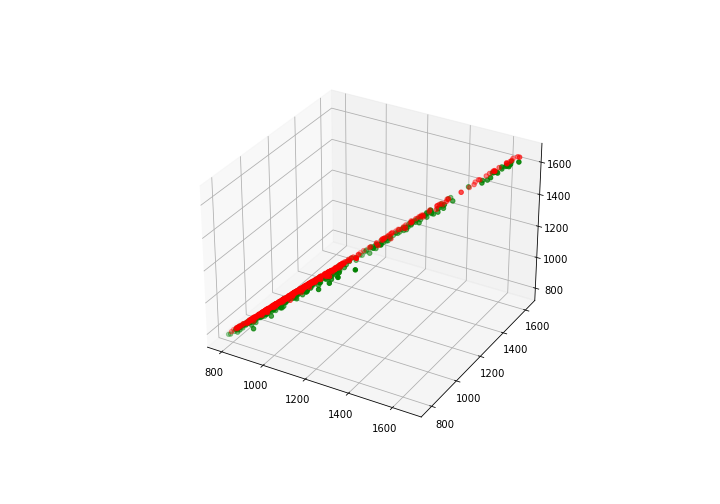}
    \caption{}
    \end{subfigure}
    \caption{Plot of four original OHLC features with respect to one another. (Green points represent Bullish days and red points represent Bearish days.) (a) Features O, H and L, (b) Features O, H and C, (c) Features O, L and C, (d) Features H, L and C.}
    \label{fig:FOHLC}
\end{figure}
\begin{figure}[htp]
    \centering
    \begin{subfigure}{0.45\textwidth}
    \centering
    \includegraphics[width=1\linewidth]{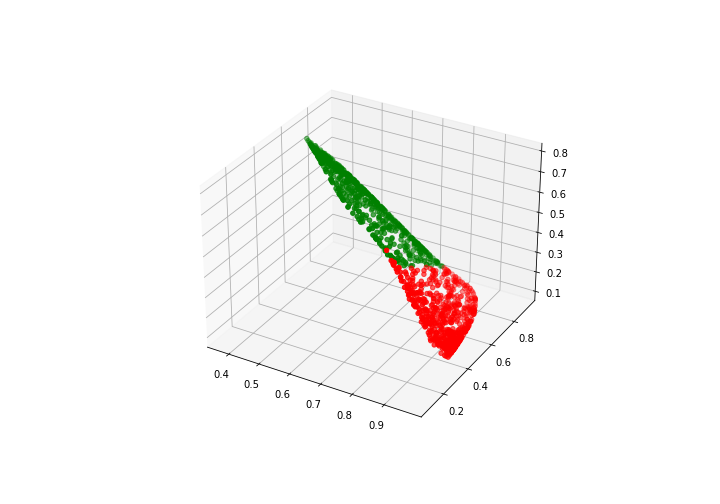}
    \caption{}
    \end{subfigure}
    \begin{subfigure}{0.45\textwidth}
    \centering
    \includegraphics[width=1\linewidth]{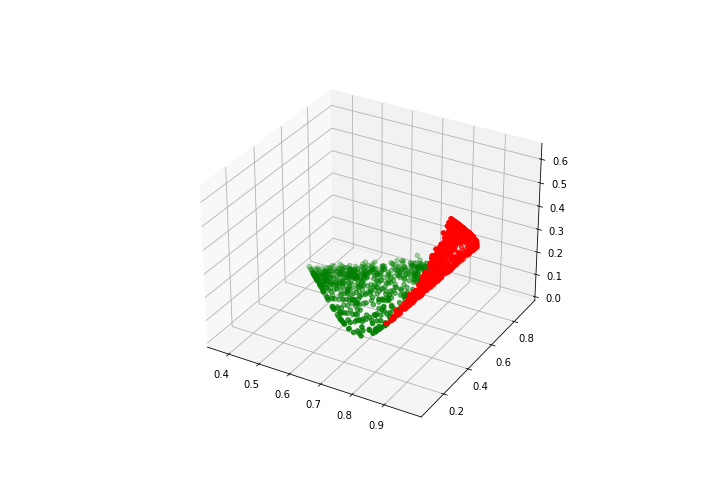}
    \caption{}
    \end{subfigure}
    \begin{subfigure}{0.45\textwidth}
    \centering
    \includegraphics[width=1\linewidth]{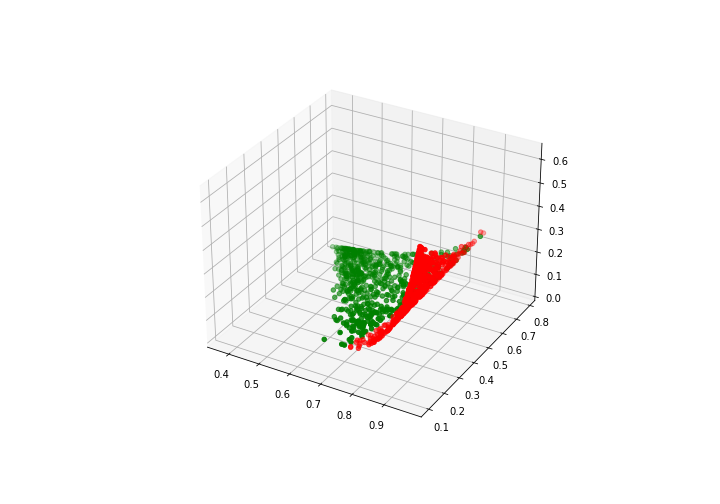}
    \caption{}
    \end{subfigure}
    \begin{subfigure}{0.45\textwidth}
    \centering
    \includegraphics[width=1\linewidth]{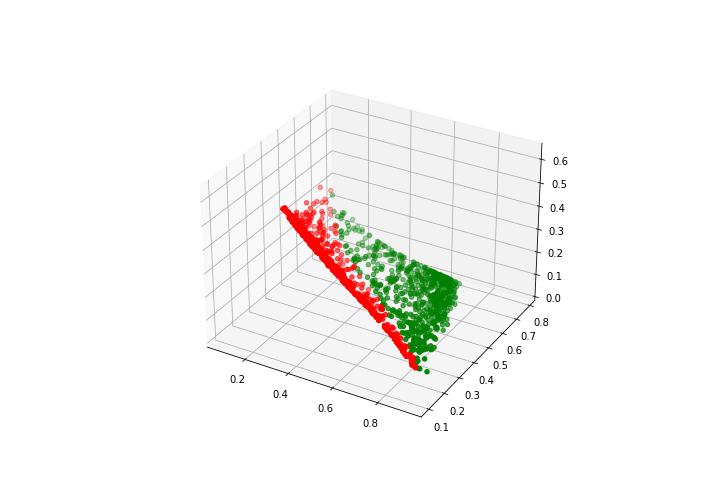}
    \caption{}
    \end{subfigure}
    \caption{Plot of four AE extracted features with respect to one another. (Green points represent Bullish days and red points represent Bearish days.) (a) Features 1, 2 and 3, (b) Features 1, 2 and 4, (c) Features 1, 3 and 4, (d) Features 2, 3 and 4.}
    \label{fig:FAE}
\end{figure}
The original OHLC feature space only captured the difference between values, but the interesting relationships between the prices were completely lost. As seen in figure \ref{fig:FOHLC}, none of the four OHLC features could show any difference between the bull or bear days. However, AE was not so overwhelmed with the difference in prices and the four features extracted by AE could easily differentiate between the bull and bear days (Figure \ref{fig:FAE}).
\subsection{Existing Methods Fail to Capture OHLC Constraints}
The proposed method used rolling windows model for all the RNN based models. The predicted prices are a stock's open price, close price, high price, low price for the next day. In all the cases a separate network is trained for each company in the dataset. Since the existing models used for stock price prediction do not take into account the OHLC constraints, we have compared the proposed method with only those models which predicted proper OHLC prices to keep the comparison fair. Although some existing methods \cite{mootha2020stock, manurung2018algorithm, wadi2018predicting}claim to predict OHLC, they fail to capture the constraints. In order to support our claim that existing methods fail to follow OHLC constraints, we have tested checked the predictions for all the 27 companies. If any of the constraints do not hold for any of the test days, we have increased the number of failure counts by one. The observations are given in Table \ref{tab:fail}.
\begin{table}[htp]
\resizebox{\textwidth}{!}
{
\begin{tabular}{lllll}
\hline\hline
Name of Stock       & Proposed & ARIMA\cite{wadi2018predicting} & MTL with BI-LSTM \cite{mootha2020stock} & LSTM \cite{manurung2018algorithm} \\\hline\hline
APSEZ               & 0        & 1     & 0                & 1    \\
ASIAN PAINTS        & 0        & 0     & 1                & 0    \\
AXIS BANK           & 0        & 2     & 1                & 2    \\
BAJAJ AUTO          & 0        & 2     & 15               & 10   \\
BAJAJ FINSERV       & 1        & 3     & 20               & 23   \\
BPCL                & 0        & 1     & 1                & 2    \\
AIRTEL              & 0        & 0     & 0                & 0    \\
CIL                 & 0        & 1     & 2                & 1    \\
GAIL INDIA          & 0        & 2     & 0                & 0    \\
HCL                 & 0        & 2     & 2                & 3    \\
HDFC BANK           & 0        & 1     & 2                & 7    \\
HINDALCO            & 0        & 2     & 0                & 2    \\
HP                  & 0        & 0     & 1                & 1    \\
ICICI BANK          & 0        & 4     & 18               & 12   \\
ITC                 & 0        & 0     & 0                & 4    \\
IOCL                & 0        & 1     & 9                & 12   \\
INFOSYS             & 0        & 1     & 0                & 1    \\
KOTAK MAHINDRA BANK & 0        & 0     & 0                & 0    \\
L\&T                & 0        & 1     & 3                & 2    \\
M  \& M             & 0        & 0     & 0                & 1    \\
NTPC                & 0        & 0     & 0                & 0    \\
ONGC                & 0        & 2     & 0                & 0    \\
PFC                 & 0        & 0     & 0                & 0    \\
POWERGRID           & 0        & 0     & 0                & 0    \\
PNB                 & 0        & 2     & 14               & 9    \\
REL                 & 0        & 0     & 5                & 17   \\
RELIANCE            & 0        & 3     & 2                & 13   \\\hline\hline
\end{tabular}}
\caption{Number of test days when the existing methods fail to follow the OHLC constraints.}
\label{tab:fail}
\end{table}

It can be seen that the proposed method maintains the OHLC constraints for all the companies' stocks for all the test days while the existing methods fail for multiple companies.
\subsection{Comparison of Methods}
It would be unfair to compare the proposed model with the methods which fail to follow OHLC constraints. Thus, standard models like ARIMA \cite{wadi2018predicting}, VAR (vector auto-regression) \cite{suharsono2017comparison}, SVR (support vector regression) \cite{henrique2018stock}, VEC (vector error correction) \cite{suharsono2017comparison}, etc which have no intrinsic ability to capture the OHLC constraints, were not considered. However, there are some methods which capture the OHLC constraints like the FLF-LSTM \cite{ahmed2020flf} (even though they explicitly do not handle it) and existing methods with LSTM \cite{manurung2018algorithm} and Bi-LSTM based multi-task model \cite{mootha2020stock} may be modified according to the proposed features and AE to predict the proposed outputs. Hence, the following methods have been compared:
\begin{enumerate}[(i)]
\item Proposed AE-MTL hybrid model with the proposed features as inputs. Let us denote it by PF-AE-PMTL.
\item Proposed MTL model without AE and just the proposed features as inputs. Let us denote it by PF-PMTL.
\item LSTM model \cite{manurung2018algorithm} without AE and just the proposed features as inputs. Let us denote it by PF-LSTM.
\item LSTM model \cite{manurung2018algorithm} with AE and the proposed features as inputs. Let us denote it by PF-AE-LSTM.
\item Bi-LSTM based multi-task model \cite{mootha2020stock} without AE and just the proposed features as inputs. Let us denote it by PF-BiLSTM-MTL. (This model cannot be hybridised with AE as the authors have assumed the initial layers to be task specific layers.)
\item FLF-LSTM \cite{ahmed2020flf} with the original non-normalised OHLC features (as the authors suggest that the method needs no normalisation.)
\end{enumerate}

\begin{figure}[H]
    \centering
    \includegraphics[width=0.9\linewidth]{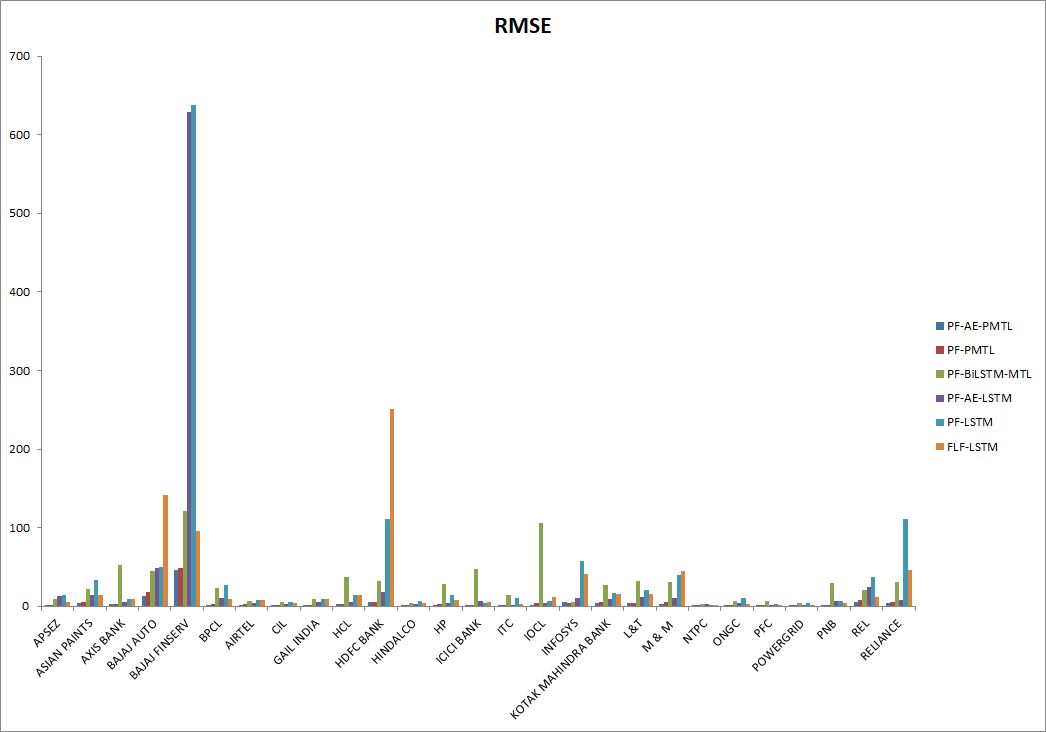}
    \caption{Comparison of root-mean squared error (RMSE) of different methods.}
    \label{fig:RMSE}
\end{figure}

\begin{figure}[H]
    \centering
    \includegraphics[width=0.9\linewidth]{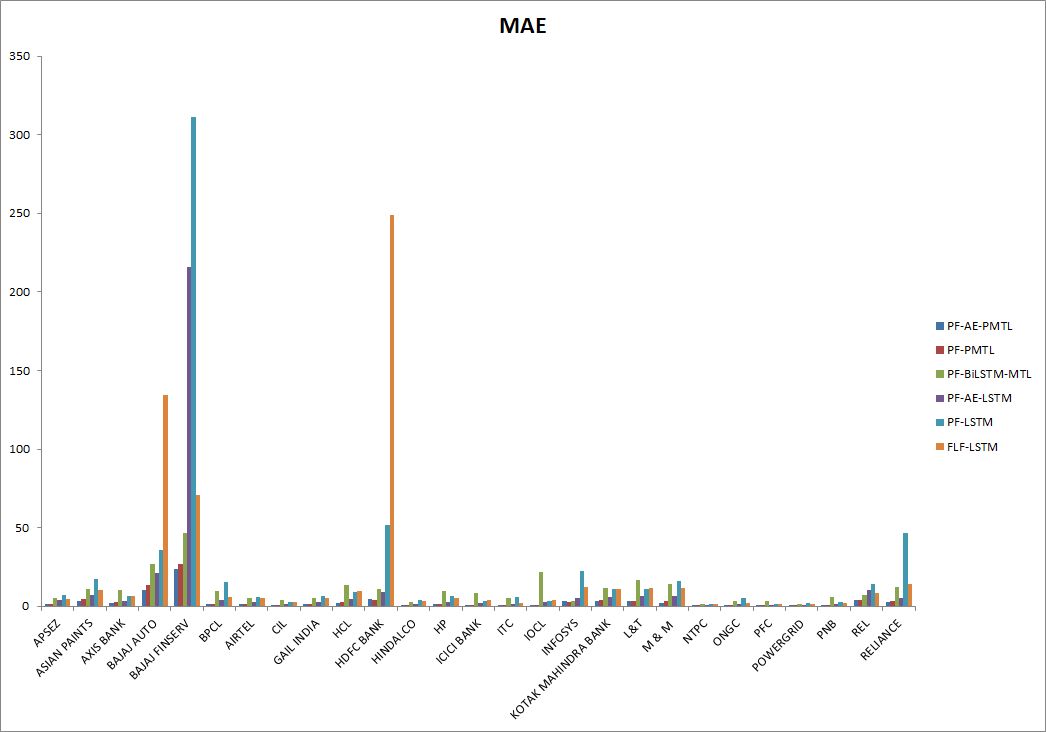}
    \caption{Comparison of mean absolute error (MAE) of different methods.}
    \label{fig:MAE}
\end{figure}

\begin{figure}[H]
    \centering
    \includegraphics[width=0.9\linewidth]{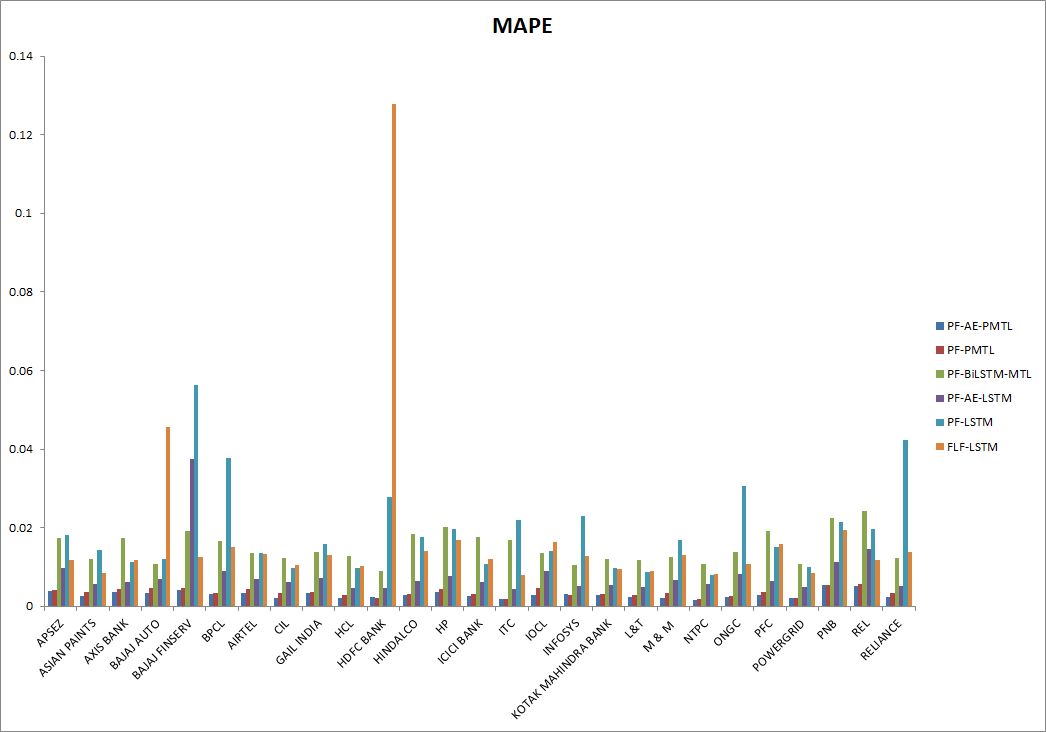}
    \caption{Comparison of mean absolute percentage error (MAPE) of different methods.}
    \label{fig:MAPE}
\end{figure}

\begin{figure}[H]
    \centering
    \includegraphics[width=0.9\linewidth]{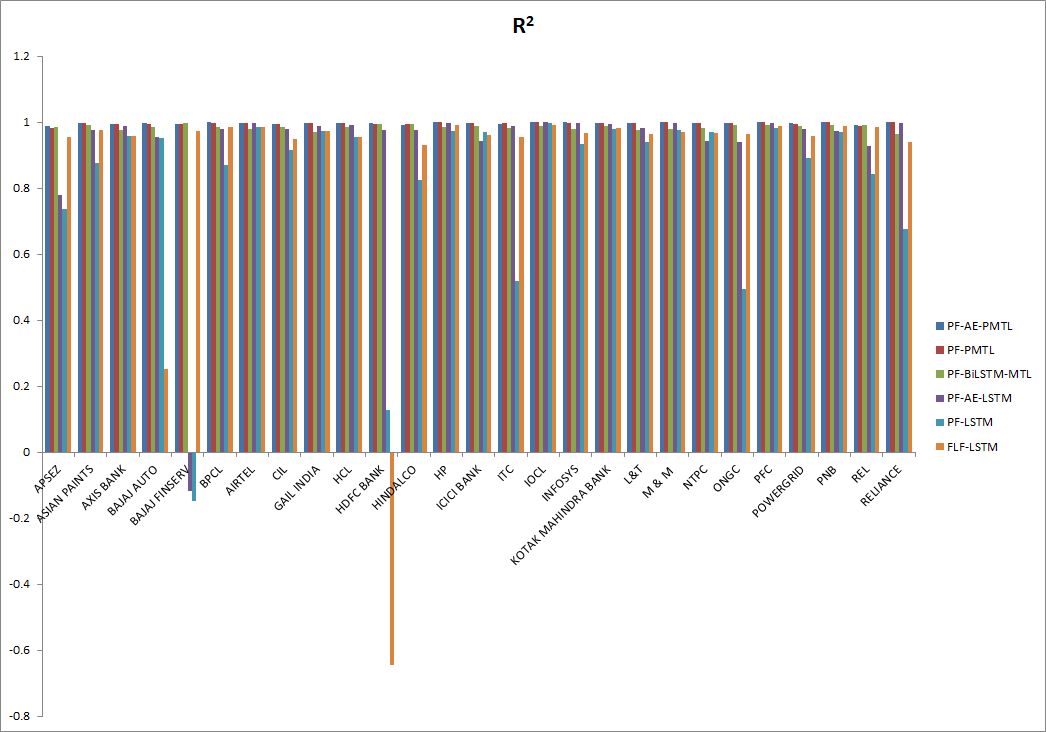}
    \caption{Comparison of R-squared ($R^2$) of different methods.}
    \label{fig:R2}
\end{figure}

Figures \ref{fig:RMSE} \ref{fig:MAE}, \ref{fig:MAPE} and \ref{fig:R2} show the root-mean square error (RMSE), mean absolute error (MAE), mean absolute percentage error (MAPE) and R-squared ($R^2$) value of different methods for various company's stocks averaged over 10 independent runs.


Let us assume that the original OHLC prices for the $i^{th}$ day are denoted by $X_O^{Actual}(i)$, $X_H^{Actual}(i)$, $X_L^{Actual}(i)$ and $X_C^{Actual}(i)$, respectively and the predicted OHLC prices for the same $i^{th}$ day are denoted by $X_O^{Pred}(i)$, $X_H^{Pred}(i)$, $X_L^{Pred}(i)$ and $X_C^{Pred}(i)$, respectively. Calculation for RMSE, MAE, MAPE and $R^2$ can be given by equations \ref{eqnrmse}, \ref{eqnmae}, \ref{eqnmape} amd \ref{eqnr2}.
\begin{multline}\label{eqnrmse}
  RMSE = \dfrac{1}{N} \sum\limits_{i=1}^N [(X_O^{Actual}(i)-X_O^{Pred}(i))^2 + (X_H^{Actual}(i)-X_H^{Pred}(i))^2 +  \\
   (X_L^{Actual}(i)-X_L^{Pred}(i))^2 + (X_C^{Actual}(i)-X_C^{Pred}(i))^2]^{\dfrac{1}{2}}.
\end{multline}
\begin{multline}\label{eqnmae}
  MAE = \dfrac{1}{N} \sum\limits_{i=1}^N [|X_O^{Actual}(i)-X_O^{Pred}(i)| + |X_H^{Actual}(i)-X_H^{Pred}(i)| +  \\
   |X_L^{Actual}(i)-X_L^{Pred}(i)| + |X_C^{Actual}(i)-X_C^{Pred}(i)|].
\end{multline}
\begin{multline}\label{eqnmape}
  MAPE = \dfrac{1}{N} \sum\limits_{i=1}^N [|\dfrac{X_O^{Actual}(i)-X_O^{Pred}(i)}{X_O^{Actual}(i)}| + |\dfrac{X_H^{Actual}(i)-X_H^{Pred}(i)}{X_H^{Actual}(i)}| +  \\
   |\dfrac{X_L^{Actual}(i)-X_L^{Pred}(i)}{X_L^{Actual}(i)}| + |\dfrac{X_C^{Actual}(i)-X_C^{Pred}(i)}{X_C^{Actual}(i)}|].
\end{multline}
\begin{equation}\label{eqnr2}
  R^2 = 1-\dfrac{SS_{Residual}}{SS_{Total}},
\end{equation}
where, the sum of squared residuals ($SS_{Residual}$) and total squared error ($SS_{Total}$) are given by,
\begin{multline}\label{eqnr21}
  SS_{Residual} = \sum\limits_{i=1}^N [(X_O^{Actual}(i)-X_O^{Pred}(i)^2 + (X_H^{Actual}(i)-X_H^{Pred}(i))^2 +  \\
   (X_L^{Actual}(i)-X_L^{Pred}(i))^2 + (X_C^{Actual}(i)-X_C^{Pred}(i))^2],
\end{multline}
and,
\begin{multline}\label{eqnr22}
  SS_{Total} = \sum\limits_{i=1}^N [(X_O^{Actual}(i)-mean(X_O^{Actual}))^2 + (X_H^{Actual}(i)-mean(X_H^{Actual}))^2 +  \\
   (X_L^{Actual}(i)-mean(X_L^{Actual}))^2 + (X_C^{Actual}(i)-mean(X_C^{Actual}))^2].
\end{multline}

It can be seen that the proposed method (PF-AE-PMTL) has the lowest RMSE, MAE and MAPE for all the stocks. The proposed method has also consistently achieved a $R^2$ value close to 1 for all the stocks (a model that has $R^2$ value equal to 1 is a model with the best performance). Some of the methods have shown negative $R^2$ values indicating that the model does not capture the trend of the data at all. The results also show that the use of AE has clearly enhanced the performance of the proposed MTL model. It is evident from the fact that the RMSE, MAE and MAPE for PF-AE-PMTL is lesser than that of PF-PMTL for all the stocks. This shows that the proposed model is comparatively a better model than the other models in terms of predicting the stock prices accurately.
\begin{figure}[H]
    \centering
    \includegraphics[width=0.9\linewidth]{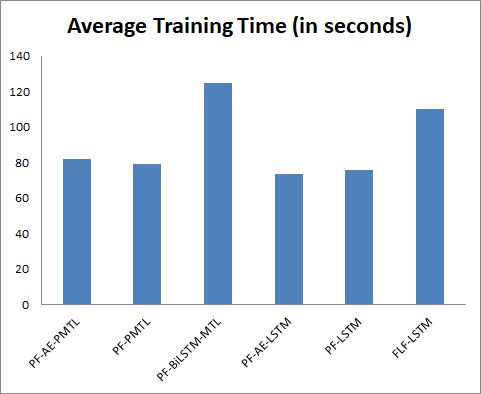}
    \caption{Comparison of the average training time of different methods.}
    \label{fig:time}
\end{figure}
The average training time taken for training of each method is shown in Figure \ref{fig:time}. The values are averaged over 10 independent runs of each method. It can be seen that the training time for the proposed method is comparable to the method with the lowest time for training (PF-AE-LSTM) and differs only by around 8 seconds. This shows that the proposed method is not so expensive in terms of training time.
\subsection{Look-back Period or Window Size}\label{lb}
In order to know which look-back period was sufficient for training the proposed model, a validation data (10\%) was separated out of the training days. This consisted of the last 123 days of the training days and the various parameters of the proposed model was validated on this particular chunk of the data.
\begin{figure}[htp]
    \centering
    \begin{subfigure}{0.3\textwidth}
    \centering
    \includegraphics[width=0.9\linewidth]{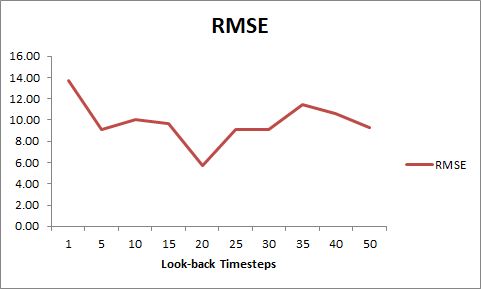}
    \caption{}
    \end{subfigure}
    \begin{subfigure}{0.3\textwidth}
    \centering
    \includegraphics[width=0.9\linewidth]{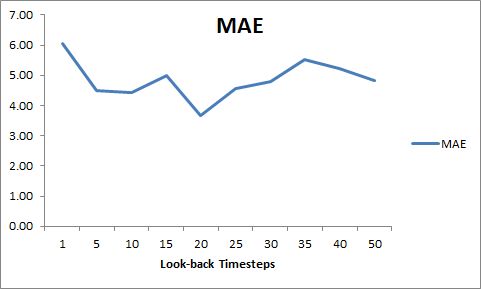}
    \caption{}
    \end{subfigure}
    \begin{subfigure}{0.3\textwidth}
    \centering
    \includegraphics[width=0.9\linewidth]{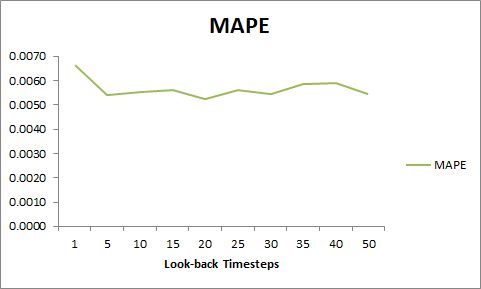}
    \caption{}
    \end{subfigure}
    \caption{Plot of the (a)RMSE, (b)MAE and (c)MAPE for various look-back periods.}
    \label{fig:lookback}
\end{figure}
The RMSE, MAE and MAPE were calculated for 10 independent runs of the model for various look-back periods upto 50 days for the Reliance India Limited's stock. It can be seen [Figure \ref{fig:lookback}] that the look-back period of 20 days was the best in terms of all the three metrics. Thus, the look-back period for 20 was chosen for the proposed method. Increasing the look-back period might have resulted in lesser error rates, but increasing the look-back period would imply a higher dimensional input and hence increased complexity.
\subsection{Number of Shared and Task-Specific Layers}
Finding the number of shared and task-specific layers is a crucial task for the MTL models in general. Thus, various experiments were conducted to fix the architecture of the proposed MTL model. Since the model already had an encoder layer and a final dense layer is needed for the output, the number of shared and task-specific layers indicate the hidden layers in between. The experiments were conducted on the validation data as used in section \ref{lb}.
\begin{table}[htp]
\centering
\begin{tabular}{llll}
\hline\hline
\textbf{Number of Shared Layers} & \textbf{RMSE} & \textbf{MAE} & \textbf{MAPE} \\\hline\hline
0                                & 24.65        & 15.29       & 0.0133        \\
1                                & 15.38        & 5.51        & 0.0070        \\
2                                & 5.72         & 3.66        & 0.0052        \\
3                                & 10.29        & 5.11        & 0.0057        \\
4                                & 7.03         & 4.46        & 0.0054        \\
5                                & 12.81        & 4.6        & 0.0058        \\\hline\hline
\end{tabular}
\caption{The RMSE, MAE and MAPE values for different number of shared layers in the model for the Reliance India Limited's stock.}
\label{tab_shared}
\end{table}
\begin{table}[htp]
\centering
\begin{tabular}{llll}
\hline\hline
\textbf{Number of task-specific layers} & \textbf{RMSE} & \textbf{MAE} & \textbf{MAPE} \\\hline\hline
0                                       & 258.77        & 162.78       & 0.1498        \\
1                                       & 5.72          & 3.66         & 0.0052        \\
2                                       & 17.46         & 4.49         & 0.0076        \\
3                                       & 9.80          & 4.05         & 0.0066        \\
4                                       & 21.08         & 4.99         & 0.0094        \\
5                                       & 9.08          & 3.79         & 0.0062        \\\hline\hline
\end{tabular}
\caption{The RMSE, MAE and MAPE values for different number of task-specific layers in the model for the Reliance India Limited's stock.}
\label{tab_tsl}
\end{table}
It can be seen from tables \ref{tab_shared} and \ref{tab_tsl} that having two shared layers and one task-specific layers resulted in the least error. It is to be noted that the This is analogous to the fact that too much depth in a DNN is often not good.
\subsection{Recommender system}
\begin{figure}[H]
    \centering
    \includegraphics[width=0.9\linewidth]{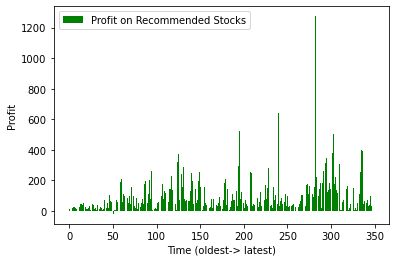}
    \caption{Actual profit obtained by the recommendations of the model PF-LSTM.}
    \label{fig:PF-LSTM}
\end{figure}

\begin{figure}[H]
    \centering
    \includegraphics[width=0.9\linewidth]{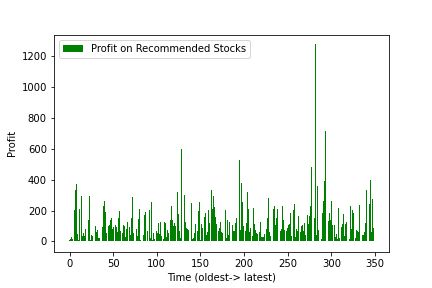}
    \caption{Actual profit obtained by the recommendations of the model PF-AE-LSTM.}
    \label{fig:PF-AE-LSTM}
\end{figure}

\begin{figure}[H]
    \centering
    \includegraphics[width=0.9\linewidth]{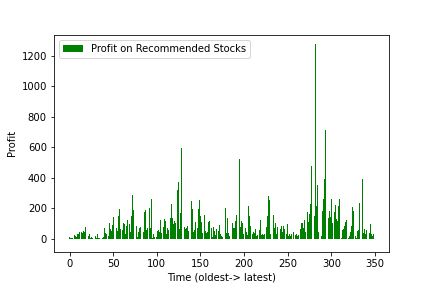}
    \caption{Actual profit obtained by the recommendations of the model PF-BiLSTM-MTL.}
    \label{fig:PFBiLSTMMTL}
\end{figure}

\begin{figure}[H]
    \centering
    \includegraphics[width=0.9\linewidth]{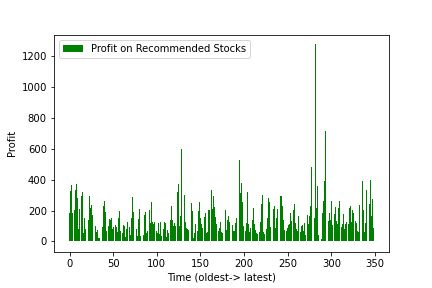}
    \caption{Actual profit obtained by the recommendations of the model FLF-LSTM.}
    \label{fig:FLF-LSTM}
\end{figure}

\begin{figure}[H]
    \centering
    \includegraphics[width=0.9\linewidth]{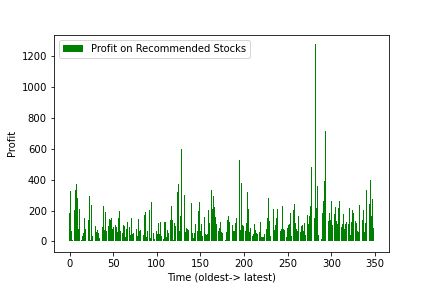}
    \caption{Actual profit obtained by the recommendations of the model PF-PMTL.}
    \label{fig:PF-PMTL}
\end{figure}

\begin{figure}[H]
    \centering
    \includegraphics[width=0.9\linewidth]{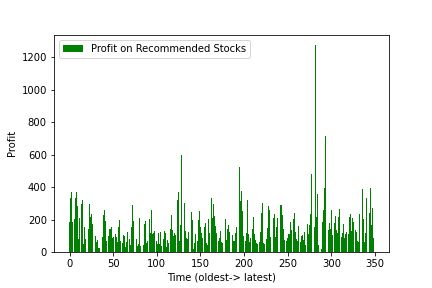}
    \caption{Actual profit obtained by the recommendations of the proposed model PF-AE-PMTL.}
    \label{fig:PF-AE-PMTL}
\end{figure}

In order to show the applicability of the proposed model as a deployable solution, the proposed model has been used to recommend stocks which would prove to be the most profitable next day. In order to do so, $Profit$ was calculated as given in equation \ref{eqn101} and based on that the top profitable stocks were recommended. The recommender system would predict the high price for the next day and would recommend the stocks for which the net profit per share is the maximum. Since the $Profit$ (from equation \ref{eqn101}) is a predicted value and holds no practical relevance, to show the efficacy of the model, the $Actual Profit$ (as calculated by equation \ref{eqn10}), was plotted against each of the test days [Figures \ref{fig:PF-LSTM} - \ref{fig:PF-AE-PMTL} ]. The intuition is that a good recommendation model will cause lesser (or almost no) losses on any of the test days. It can be seen from Figures \ref{fig:PF-LSTM}, \ref{fig:PF-AE-LSTM}, \ref{fig:PFBiLSTMMTL} and \ref{fig:PF-PMTL} that all the other methods had multiple occasions in the test days where the $Actual Profit$ was negative (which indicates a loss). However, as seen from Figure \ref{fig:PF-AE-PMTL}, the proposed model had not made a loss for even a single day for all the test days.
The next recommendation is based on William \%R. This is a very important trend indicator for the investors. It is said that the investors usually buy the stock when the William \%R is  below -80. In order to give this particular type of recommendation, the proposed PF-AE-PMTL was analysed for ten random test dates.

\begin{table}[htp]
\begin{center}
\begin{tabular}{llll}
\hline\hline
\textbf{Date} & \textbf{Stocks} & \textbf{Predicted \%R} & \textbf{Actual \%R} \\\hline\hline
14-09-2017 & RELIANCE      & 98.39 & 96.26 \\\hline
10-10-2017 & L\&T          & 80.35 & 88.09 \\\hline
15-11-2017 & IOCL          & 81.04 & 92.94 \\\hline
22-02-2018 & BAJAJ AUTO    & 82.73 & 97.15 \\
           & IOCL          & 81.41 & 96.61 \\
           & PNB           & 87.41 & 94.26 \\\hline
21-03-2018 & IOCL          & 93.60 & 99.17 \\
           & PFC           & 83.83 & 78.88 \\\hline
23-03-2018 & IOCL          & 95.37 & 98.70 \\\hline
25-04-2018 & BPCL          & 80.76 & 81.34 \\\hline
           & HP            & 82.72 & 98.49 \\
           & PNB           & 80.16 & 97.97 \\\hline
10-05-2018 & CIL           & 80.67 & 88.55 \\
           & HCL           & 87.00 & 88.97 \\\hline
28-05-2018 & RELIANCE      & 83.89 & 86.76 \\\hline
10-09-2018 & BAJAJ FINSERV & 80.19 & 94.50 \\
           & INFOSYS       & 97.04 & 98.01 \\\hline\hline
\end{tabular}
\end{center}
\caption{Comparison of actual and predicted William's \%R for stocks which were recommended by the proposed model for buying.}
\label{tab_inwill}
\end{table}

It can be seen that most of the recommended stocks actually crossed the -80 mark of William \%R on the five random test days. Only one recommended stock (PFC) did not cross the -80 mark on 21-03-2018, but the actual value of its William's \%R was close to -80.

It can be seen from the experiments that the proposed model is quite efficient in making profitable recommendations to the users who follow the predictions for investing in stocks.

\section{Conclusions and Future Work}\label{C5sec6}
This manuscript presents an hybrid AE multi-task model for stock prediction. The work has been extended to build a recommender system. The inclusion of pre-trained encoder modules clearly helped in increasing the predictive capabilities of the proposed MTL model for OHLC predictions. The novelty of the proposed model lies in the unique normalisation strategy for OHLC data and explains why an AE might be better to use in problems like stock markets where the prices may fluctuate drastically. This angle of stock price prediction has mostly been overlooked in the literature. Thus, this work tries to address the gap of using ML in financial prediction tasks.

It is to be noted that the proposed work considers straightforward price values of stocks and does not consider any indicator variables used by the financial analysts. However, we are hoping to extend this work in future using other technical indicators as well. In this present analysis the information regarding the volume of stocks sold is not incorporated as the data for some companies had no information on volume. In future, we therefore wish to incorporate some volume indicators as well. We also hope to make a composite model in future which would combine both the fundamental as well as technical analysis for recommendation of stocks. The proposed models would have been more practical if the predictions could have given for very short-term or very long-term trading purposes, which is also one of the planned future directions. 

%% file: Main.bbl
\begin{thebibliography}{10}

\bibitem{abu1996introduction}
Y.~S Abu-Mostafa and Amir~F Atiya.
\newblock Introduction to {F}inancial {F}orecasting.
\newblock {\em Applied Intelligence}, 6(3):205--213, 1996.

\bibitem{adam2016stock}
K.~Adam, A.~Marcet, and J.~P. Nicolini.
\newblock Stock {M}arket {V}olatility and {L}earning.
\newblock {\em The Journal of Finance}, 71(1):33--82, 2016.

\bibitem{adebiyi2014comparison}
A.~A. Adebiyi, A.~O. Adewumi, and C.~K. Ayo.
\newblock Comparison of {A}{R}{I}{M}{A} and {A}rtificial {N}eural {N}etworks
  {M}odels for {S}tock {P}rice {P}rediction.
\newblock {\em Journal of Applied Mathematics}, 2014, 2014.

\bibitem{ahmed2020flf}
S.~Ahmed, S.~U. Hassan, N.~R. Aljohani, and R.~Nawaz.
\newblock {F}{L}{F}-{L}{S}{T}{M}: {A} {N}ovel {P}rediction {S}ystem using
  {F}orex {L}oss {F}unction.
\newblock {\em Applied Soft Computing}, 97:106780, 2020.

\bibitem{ballings2015evaluating}
M.~Ballings, D.~Van~den Poel, N.~Hespeels, and R.~Gryp.
\newblock Evaluating {M}ultiple {C}lassifiers for {S}tock {P}rice {D}irection
  {P}rediction.
\newblock {\em Expert Systems with Applications}, 42(20):7046--7056, 2015.

\bibitem{baxter1997bayesian}
J.~Baxter.
\newblock A {B}ayesian/ {I}nformation {T}heoretic {M}odel of {L}earning to
  {L}earn via {M}ultiple {T}ask {S}ampling.
\newblock {\em Machine learning}, 28(1):7--39, 1997.

\bibitem{bhanja2019impact}
S.~Bhanja and A.~Das.
\newblock Impact of {D}ata {N}ormalization on {D}eep {N}eural {N}etwork for
  {T}ime {S}eries {F}orecasting.
\newblock In {\em Proceedings of Conference on Advancement in Computation,
  Communication and Electronics Paradigm (ACCEP-2019)}, page~27, 2019.

\bibitem{budiharto2021data}
Widodo Budiharto.
\newblock Data {S}cience {A}pproach to {S}tock {P}rices {F}orecasting in
  {I}ndonesia during {C}ovid-19 using {L}ong {S}hort-{T}erm {M}emory
  ({L}{S}{T}{M}).
\newblock {\em Journal of big data}, 8(1):1--9, 2021.

\bibitem{chakraborty2019integration}
D.~Chakraborty, V.~Narayanan, and A.~Ghosh.
\newblock Integration of {D}eep {F}eature {E}xtraction and {E}nsemble
  {L}earning for {O}utlier {D}etection.
\newblock {\em Pattern Recognition}, 89:161--171, 2019.

\bibitem{chen2015hybrid}
M.~Y. Chen and B.~T. Chen.
\newblock A {H}ybrid {F}uzzy {T}ime {S}eries {M}odel based on {G}ranular
  {C}omputing for {S}tock {P}rice {F}orecasting.
\newblock {\em Information Sciences}, 294:227--241, 2015.

\bibitem{chen2017enhancement}
Y.~J. Chen, Y.~M. Chen, and C.~L. Lu.
\newblock Enhancement of {S}tock {M}arket {F}orecasting using an {I}mproved
  {F}undamental {A}nalysis-based {A}pproach.
\newblock {\em Soft Computing}, 21(13):3735--3757, 2017.

\bibitem{colby2003encyclopedia}
R.~W. Colby.
\newblock {\em The {E}ncyclopedia of {T}echnical {M}arket {I}ndicators,
  {S}econd {E}dition}.
\newblock McGraw-Hill Education, 2003.

\bibitem{dash2016hybrid}
R.~Dash and P.~K. Dash.
\newblock A {H}ybrid {S}tock {T}rading {F}ramework {I}ntegrating {T}echnical
  {A}nalysis with {M}achine {L}earning {T}echniques.
\newblock {\em The Journal of Finance and Data Science}, 2(1):42--57, 2016.

\bibitem{de2017complexity}
K.~De~Bot.
\newblock Complexity {T}heory and {D}ynamic {S}ystems {T}heory.
\newblock {\em Complexity Theory and Language Development: In Celebration of
  Diane Larsen-Freeman}, 48:51, 2017.

\bibitem{edwards2018technical}
R.~D. Edwards, J.~Magee, and W.~H.~C. Bassetti.
\newblock {\em Technical {A}nalysis of {S}tock {T}rends}.
\newblock Taylor \& Francis, 2018.

\bibitem{engle2013stock}
R.~F. Engle, E.~Ghysels, and B.~Sohn.
\newblock Stock {M}arket {V}olatility and {M}acroeconomic {F}undamentals.
\newblock {\em Review of Economics and Statistics}, 95(3):776--797, 2013.

\bibitem{enke2013stock}
D.~Enke and N.~Mehdiyev.
\newblock Stock {M}arket {P}rediction using a {C}ombination of {S}tepwise
  {R}egression {A}nalysis, {D}ifferential {E}volution-based {F}uzzy
  {C}lustering, and a {F}uzzy {I}nference {N}eural {N}etwork.
\newblock {\em Intelligent Automation \& Soft Computing}, 19(4):636--648, 2013.

\bibitem{fama1965behavior}
E.~F. Fama.
\newblock The {B}ehavior of {S}tock-{M}arket {P}rices.
\newblock {\em The Journal of Business}, 38(1):34--105, 1965.

\bibitem{fischer2018deep}
T.~Fischer and C.~Krauss.
\newblock Deep {L}earning with {L}ong {S}hort-{T}erm {M}emory {N}etworks for
  {F}inancial {M}arket {P}redictions.
\newblock {\em European Journal of Operational Research}, 270(2):654--669,
  2018.

\bibitem{francq2019garch}
C.~Francq and J.~M. Zakoian.
\newblock {\em G{A}{R}{C}{H} {M}odels: {S}tructure, {S}tatistical {I}nference
  and {F}inancial {A}pplications}.
\newblock John Wiley \& Sons, 2019.

\bibitem{gorenc2016prediction}
M.~Gorenc~Novak and D.~Velu{\v{s}}{\v{c}}ek.
\newblock Prediction of {S}tock {P}rice {M}ovement based on {D}aily {H}igh
  {P}rices.
\newblock {\em Quantitative Finance}, 16(5):793--826, 2016.

\bibitem{greff2016lstm}
K.~Greff, R.~K. Srivastava, J.~Koutn{\'\i}k, B.~R. Steunebrink, and
  J.~Schmidhuber.
\newblock L{S}{T}{M}: {A} {S}earch {S}pace {O}dyssey.
\newblock {\em IEEE Transactions on Neural Networks and Learning Systems},
  28(10):2222--2232, 2016.

\bibitem{henrique2018stock}
B.~M. Henrique, V.~A. Sobreiro, and H.~Kimura.
\newblock Stock {P}rice {P}rediction using {S}upport {V}ector {R}egression on
  {D}aily and {U}p to the {M}inute {P}rices.
\newblock {\em The Journal of Finance and Data Science}, 4(3):183--201, 2018.

\bibitem{hochreiter1997long}
S.~Hochreiter and J.~Schmidhuber.
\newblock Long {S}hort-{T}erm {M}emory.
\newblock {\em Neural computation}, 9(8):1735--1780, 1997.

\bibitem{hussain2019cloud}
Tanveer Hussain, Khan Muhammad, Amin Ullah, Zehong Cao, Sung~Wook Baik, and
  Victor Hugo~C de~Albuquerque.
\newblock Cloud-assisted multiview video summarization using cnn and
  bidirectional lstm.
\newblock {\em IEEE Transactions on Industrial Informatics}, 16(1):77--86,
  2019.

\bibitem{karalevicius2018using}
V.~Karalevicius, N.~Degrande, and J.~De~Weerdt.
\newblock Using {S}entiment {A}nalysis to {P}redict {I}nterday {B}itcoin
  {P}rice {M}ovements.
\newblock {\em The Journal of Risk Finance}, 19(1):56--75, 2018.

\bibitem{kim2018forecasting}
H.~Y. Kim and C.~H. Won.
\newblock Forecasting the {V}olatility of {S}tock {P}rice {I}ndex: {A} {H}ybrid
  {M}odel {I}ntegrating {L}{S}{T}{M} with {M}ultiple {G}{A}{R}{C}{H}-type
  {M}odels.
\newblock {\em Expert Systems with Applications}, 103:25--37, 2018.

\bibitem{malkiel2003efficient}
B.~G. Malkiel.
\newblock The {E}fficient {M}arket {H}ypothesis and its {C}ritics.
\newblock {\em Journal of Economic Perspectives}, 17(1):59--82, 2003.

\bibitem{manurung2018algorithm}
A.~H. Manurung, W.~Budiharto, and H.~Prabowo.
\newblock Algorithm and {M}odeling of {S}tock {P}rices {F}orecasting based on
  {L}ong {S}hort-{T}erm {M}emory ({L}{S}{T}{M}).
\newblock {\em International Journal of Innovative Computing Information and
  Control (ICIC)}, 12(12):1277--1283, 2018.

\bibitem{meng2018research}
L.~Meng, S.~Ding, N.~Zhang, and J.~Zhang.
\newblock Research of {S}tacked {D}enoising {S}parse {A}utoencoder.
\newblock {\em Neural Computing and Applications}, 30(7):2083--2100, 2018.

\bibitem{mondal2014study}
P.~Mondal, L.~Shit, and S.~Goswami.
\newblock Study of {E}ffectiveness of {T}ime {S}eries modeling
  ({A}{R}{I}{M}{A}) in {F}orecasting {S}tock {P}rices.
\newblock {\em International Journal of Computer Science, Engineering and
  Applications}, 4(2):13, 2014.

\bibitem{mootha2020stock}
S.~Mootha, S.~Sridhar, R.~Seetharaman, and S.~Chitrakala.
\newblock Stock {P}rice {P}rediction using {B}i-{D}irectional {L}{S}{T}{M}
  based {S}equence to {S}equence {M}odeling and {M}ultitask {L}earning.
\newblock In {\em 2020 11th IEEE Annual Ubiquitous Computing, Electronics \&
  Mobile Communication Conference (UEMCON)}, pages 0078--0086. IEEE, 2020.

\bibitem{nayak2015naive}
R.~K. Nayak, D.~Mishra, and A.~K. Rath.
\newblock A {N}a{\"\i}ve {S}{V}{M}-$k${N}{N} based {S}tock {M}arket {T}rend
  {R}eversal {A}nalysis for {I}ndian {B}enchmark {I}ndices.
\newblock {\em Applied Soft Computing}, 35:670--680, 2015.

\bibitem{pan2009survey}
S.~J. Pan and Q.~Yang.
\newblock A {S}urvey on {T}ransfer {L}earning.
\newblock {\em IEEE Transactions on Knowledge and Data Engineering},
  22(10):1345--1359, 2009.

\bibitem{panigrahi2013effect}
S.~Panigrahi and H.~S. Behera.
\newblock Effect of {N}ormalization {T}echniques on {U}nivariate {T}ime
  {S}eries {F}orecasting using {E}volutionary {H}igher {O}rder {N}eural
  {N}etwork.
\newblock {\em International Journal of Engineering and Advanced Technology},
  3(2):280--285, 2013.

\bibitem{reyes2019performing}
O.~Reyes and S.~Ventura.
\newblock Performing {M}ulti-{T}arget {R}egression via a {P}arameter
  {S}haring-based {D}eep {N}etwork.
\newblock {\em International journal of neural systems}, 29(09):1950014, 2019.

\bibitem{shen2018deep}
G.~Shen, Q.~Tan, H.~Zhang, P.~Zeng, and J.~Xu.
\newblock Deep {L}earning with {G}ated {R}ecurrent {U}nit {N}etworks for
  {F}inancial {S}equence {P}redictions.
\newblock {\em Procedia computer science}, 131:895--903, 2018.

\bibitem{shibata2009effect}
K.~Shibata and Y.~Ikeda.
\newblock Effect of {N}umber of {H}idden {N}eurons on {L}earning in
  {L}arge-{S}cale {L}ayered {N}eural {N}etworks.
\newblock In {\em 2009 ICCAS-SICE}, pages 5008--5013. IEEE, 2009.

\bibitem{singh2017stock}
R.~Singh and S.~Srivastava.
\newblock Stock {P}rediction using {D}eep {L}earning.
\newblock {\em Multimedia Tools and Applications}, 76(18):18569--18584, 2017.

\bibitem{sola1997importance}
J.~Sola and J.~Sevilla.
\newblock Importance of {I}nput {D}ata {N}ormalization for the {A}pplication of
  {N}eural {N}etworks to {C}omplex {I}ndustrial {P}roblems.
\newblock {\em IEEE Transactions on Nuclear Science}, 44(3):1464--1468, 1997.

\bibitem{suharsono2017comparison}
A.~Suharsono, A.~Aziza, and W.~Pramesti.
\newblock Comparison of {V}ector {A}utoregressive ({V}{A}{R}) and {V}ector
  {E}rror {C}orrection {M}odels ({V}{E}{C}{M}) for {I}ndex of {A}{S}{E}{A}{N}
  {S}tock {P}rice.
\newblock In {\em AIP Conference Proceedings}, volume 1913, page 020032. AIP
  Publishing LLC, 2017.

\bibitem{talarposhti2016stock}
F.~M. Talarposhti, H.~J. Sadaei, R.~Enayatifar, F.~G. Guimar{\~a}es, M.~Mahmud,
  and T.~Eslami.
\newblock Stock {M}arket {F}orecasting by using a {H}ybrid {M}odel of
  {E}xponential {F}uzzy {T}ime {S}eries.
\newblock {\em International Journal of Approximate Reasoning}, 70:79--98,
  2016.

\bibitem{tiwari2012comparison}
P.~Tiwari.
\newblock Comparison of {D}ifferent {N}atural {L}anguage {P}rocessing {M}odels
  and {N}eural {N}etworks for {P}redicting {S}tock {P}rices: {A} {C}ase
  {S}tudy.
\newblock {\em IEEE Transactions on Microwave Theory and Techniques}, 60(12):1,
  2012.

\bibitem{tsai2019assessing}
Y.~C. Tsai, M.~E. Wu, J.~H. Syu, C.~L. Lei, C.~S. Wu, J.~M. Ho, and C.~J. Wang.
\newblock Assessing the {P}rofitability of {T}imely {O}pening {R}ange
  {B}reakout on {I}ndex {F}utures {M}arkets.
\newblock {\em IEEE Access}, 7:32061--32071, 2019.

\bibitem{wadi2018predicting}
S.~A. Wadi, M.~Almasarweh, A.~A. Alsaraireh, and J.~Aqaba.
\newblock Predicting {C}losed {P}rice {T}ime {S}eries {D}ata using
  {A}{R}{I}{M}{A} {M}odel.
\newblock {\em Modern Applied Science}, 12(11), 2018.

\bibitem{wang2019stock}
Y.~Wang, H.~Liu, Q.~Guo, S.~Xie, and X.~Zhang.
\newblock Stock {V}olatility {P}rediction by {H}ybrid {N}eural {N}etwork.
\newblock {\em IEEE Access}, 7:154524--154534, 2019.

\bibitem{wu2020adaptive}
X.~Wu, H.~Chen, J.~Wang, L.~Troiano, V.~Loia, and H.~Fujita.
\newblock Adaptive {S}tock {T}rading {S}trategies with {D}eep {R}einforcement
  {L}earning {M}ethods.
\newblock {\em Information Sciences}, 538:142--158, 2020.

\bibitem{xie2020bioacoustic}
J.~Xie, K.~Hu, M.~Zhu, and Y.~Guo.
\newblock Bioacoustic {S}ignal {C}lassification in {C}ontinuous {R}ecordings:
  {S}yllable-{S}egmentation vs {S}liding-{W}indow.
\newblock {\em Expert Systems with Applications}, 152:113390, 2020.

\bibitem{zahedi2015application}
J.~Zahedi and M.~M. Rounaghi.
\newblock Application of {A}rtificial {N}eural {N}etwork {M}odels and
  {P}rincipal {C}omponent {A}nalysis {M}ethod in {P}redicting {S}tock {P}rices
  on {T}ehran {S}tock {E}xchange.
\newblock {\em Physica A: Statistical Mechanics and its Applications},
  438:178--187, 2015.

\bibitem{zhang2018overview}
Y.~Zhang and Q.~Yang.
\newblock An {O}verview of {M}ulti-{T}ask {L}earning.
\newblock {\em National Science Review}, 5(1):30--43, 2018.

\bibitem{zhao2017lstm}
Z.~Zhao, W.~Chen, X.~Wu, P.~C.~Y. Chen, and J.~Liu.
\newblock L{S}{T}{M} {N}etwork: {A} {D}eep {L}earning {A}pproach for
  {S}hort-{T}erm {T}raffic {F}orecast.
\newblock {\em IET Intelligent Transport Systems}, 11(2):68--75, 2017.

\bibitem{zhong2017forecasting}
X.~Zhong and D.~Enke.
\newblock Forecasting {D}aily {S}tock {M}arket {R}eturn using {D}imensionality
  {R}eduction.
\newblock {\em Expert Systems with Applications}, 67:126--139, 2017.

\end{thebibliography}
